\documentclass{ecai}
\usepackage{graphicx}
\usepackage{latexsym}

\usepackage{microtype}
\usepackage{graphicx}
\usepackage{subfigure}
\usepackage{booktabs}
\usepackage[implicit]{hyperref}

\usepackage{amsmath}
\usepackage{amssymb}
\usepackage{mathtools}
\usepackage{amsthm}

\usepackage{graphics}
\usepackage{amsmath,amssymb,amsthm,amsfonts}
\usepackage{graphicx,multirow,enumerate}
\usepackage{subfigure}
\usepackage{color}
\usepackage{caption}
\usepackage{booktabs}
\usepackage{kotex}

\usepackage{graphicx}
\usepackage{epsfig}
\usepackage{hyperref}
\usepackage{fancyhdr}
\usepackage{colortbl}
\usepackage{multicol}
\usepackage{graphicx}
\usepackage{enumitem}
\usepackage{makecell}

\usepackage{tikz}

\usetikzlibrary{shapes,decorations,arrows,calc,arrows.meta,fit,positioning}
\tikzset{
    -Latex,auto,node distance =1 cm and 1 cm,semithick,
    state/.style ={ellipse, draw, minimum width = 0.7 cm},
    point/.style = {circle, draw, inner sep=0.04cm,fill,node contents={}},
    bidirected/.style={Latex-Latex,dashed},
    el/.style = {inner sep=2pt, align=left, sloped}
}

\def\cF{{\cal F}}

\def\cL{{\cal L}}

\def\cL{{\cal L}}
\def\cM{{\cal M}}

\newcommand{\bfa}{{\bf a}}

\newcommand{\bu}{{\bf u}}

\newcommand{\bv}{{\bf v}}

\newcommand{\bz}{{\bf z}}

\newcommand{\bx}{{\bf x}}
\newcommand{\bg}{{\bf g}}

\newcommand{\mbR}{\mathbb{R}}
\newcommand{\mbE}{\mathbb{E}}

\newcommand{\indep}{\perp \!\!\! \perp}

\newcommand{\bepsilon}{\mbox{\boldmath{$\epsilon$}}}

\newcommand{\bc}{\begin{center}}
\newcommand{\ec}{\end{center}}
\newcommand{\be}{\begin{equation}}
\newcommand{\ee}{\end{equation}}
\newcommand{\ba}{\begin{array}}
\newcommand{\ea}{\end{array}}
\newcommand{\bean}{\begin{eqnarray*}}
\newcommand{\eean}{\end{eqnarray*}}
\newcommand{\bea}{\begin{eqnarray}}
\newcommand{\eea}{\end{eqnarray}}
\newcommand{\ben}{\begin{enumerate}}
\newcommand{\een}{\end{enumerate}}
\newcommand{\bed}{\begin{itemize}}
\newcommand{\eed}{\end{itemize}}
\newcommand{\bs}{\begin{slide}}
\newcommand{\es}{\end{slide}}

\newtheorem{proposition}[theorem]{Proposition}
\newtheorem{definition}[theorem]{Definition}
\newtheorem{assumption}[theorem]{Assumption}

\begin{document}

\begin{frontmatter}

\title{Causally Disentangled Generative \\
Variational AutoEncoder}

% \author[]{Keywords: Variational AutoEncoder, causal disentanglement learning, causally disentangled generation, causal disentanglement metric}

\author[A]{\fnms{SeungHwan}~\snm{AN}}
% \orcid{0000-0002-1891-1174}
\author[B]{\fnms{Kyungwoo}~\snm{Song}}
% \orcid{0000-0003-0082-4280}
\author[A]{\fnms{Jong-June}~\snm{Jeon}\thanks{Corresponding Author. Email: jj.jeon@uos.ac.kr.}} 
% use of \orcid{} is optional

\address[A]{Department of Statistics, University of Seoul, S. Korea}
\address[B]{Department of Applied Statistics, Department of Statistics and Data Science, Yonsei University, S. Korea}

\begin{abstract}
We present a new supervised learning technique for the Variational AutoEncoder (VAE) that allows it to learn a causally disentangled representation and generate causally disentangled outcomes simultaneously. We call this approach Causally Disentangled Generation (CDG). CDG is a generative model that accurately decodes an output based on a causally disentangled representation. Our research demonstrates that adding supervised regularization to the encoder alone is insufficient for achieving a generative model with CDG, even for a simple task. Therefore, we explore the necessary and sufficient conditions for achieving CDG within a specific model. Additionally, we introduce a universal metric for evaluating the causal disentanglement of a generative model. Empirical results from both image and tabular datasets support our findings.
\end{abstract}

\end{frontmatter}

\section{Introduction}
\label{sec:1}

Learning disentangled representation is a widely studied and challenging topic of VAE \cite{kingma2013auto}, and GAN \cite{goodfellow2020generative} due to its potential to enable interpretable data generation and enhance downstream task performances \cite{szabo2017challenges}. Roughly speaking, studies on disentangled representation investigate the structure of a latent space where each dimension corresponds to a ground-truth factor that generates a dataset \cite{bengio2013representation}. In early studies of disentangled representation, the ground-truth factors consisting of the latent space are assumed to be mutually independent exogenous variables \cite{higgins2016beta}. In light of the growing interest in interpretable generative models, recent research has expanded the modeling of disentangled representation by incorporating a Structural Causal Model (SCM) and a jointly dependent model for the ground-truth factors \cite{bengio2019meta, trauble2021disentangled, suter2019robustly, reddy2022causally}. 
% The importance of supervised learning for disentangled representation is raised by \cite{locatello2019challenging}, who proved that unsupervised disentanglement learning is impossible. Especially when the model includes endogenous factors of interest, the validity of the maximum likelihood method is guaranteed only when the ground-truth factors are correctly specified \cite{trauble2021disentangled}. So, the Structural Causal Model (SCM) is crucial for constructing latent space that causally aligns with the ground-truth factors \cite{yang2021causalvae, shen2022weakly}. The alignment of the latent structure is adapted by supervised encoder regularization method \cite{Locatello2020Disentangling}, which produces causally-aware generative models \cite{xu2019achieving, van2021decaf, wen2021causal}.

% 수정
The importance of supervised learning for disentangled representation is raised by \cite{locatello2019challenging}, who proved that unsupervised disentanglement learning is impossible. Especially when the model includes endogenous factors of interest, the validity of the maximum likelihood method is guaranteed only when the ground-truth factors are correctly specified \cite{trauble2021disentangled}. So, the SCM and supervised encoder regularization method \cite{Locatello2020Disentangling} are crucial for constructing latent space that causally aligns with the ground-truth factors \cite{yang2021causalvae, shen2022weakly}. Furthermore, the alignment of the latent structure is also adapted by topological generation, which produces causally-aware generative models \cite{xu2019achieving, van2021decaf, wen2021causal}.
% So, under the assumption of causally entangled ground-truth factors and the supervision setting, \cite{kocaoglu2017causalgan, shen2022weakly, yang2021causalvae} proposed image generation models where the representation is disentangled and causally entangled. These papers also numerically showed the necessity of causal structure in latent space to learn causally disentangled representation. 
% Thus, an informal definition of causally disentangled representation is 1) the representation is causally entangled as the same as the ground-truth factors, and 2) each dimension of representation represents a single ground-truth factor.

However, we found that the supervised regularization of the encoder \cite{yang2021causalvae, shen2022weakly} is insufficient for achieving the causally-aware generative model. Even when the encoder builds a causally disentangled latent space, the causality between a latent variable and the generated output may not hold due to the entangled structure of the decoder. Our research demonstrates that causally-aware generative models are necessarily able to recover the output according to the causally disentangled factors identified by the encoder. We refer to this property as Causally Disentangled Generation (CDG) and focus on the required conditions of the decoder and the causal effect of CDG. Based on the conditions, we propose a new VAE model satisfying CDG, the CDG-VAE.

%In this paper, we define the causally disentangled generation (CDG) as  a generative mechanism satisfying a property that a causally disentangled factor identified by an encoder 

% However, since the disentangled representation only depends on the encoder, it does not guarantee the causally disentangled generation (CDG), which is defined as the causally plausible data generation according to $do$-interventions on representations. Our main contribution is that we show that the CDG is achieved by 1) the causal structure of disentangled representation and 2) a sufficient condition for a decoder structure. In short, the CDG depends on both the encoder and decoder of the VAE. 

The development of CDG-VAE makes three contributions to the field of causally-aware generative models. First, we establish sufficient and necessary conditions of the decoder structure for CDG. Second, CDG-VAE can be applied to chain graphs (i.e., Partial DAGs) unlikely \cite{xu2019achieving, van2021decaf, wen2021causal} that require a completely identified directed acyclic graph (DAG) for a topological generation. Third, we propose a generalized metric measuring the degree of causally disentangled generativeness under an arbitrary DAG structure of the ground-truth factors. Our metric is derived from the necessary conditions for CDG and $do$-calculus of causal effects \cite{pearl2009causality}.

We aim to demonstrate the effectiveness of our proposed model by evaluating two distinct types of datasets, namely image and tabular data. Specifically, we show that our model can produce causally plausible counterfactual samples with both qualitative and quantitative assessments of the image dataset. Additionally, we provide evidence of the advantages attained from the causally disentangled representation of our model in terms of two downstream tasks: sample efficiency and distributional robustness \cite{shen2022weakly}. Moving on to the tabular dataset, we show that our model can generate high-quality synthetic data while preserving the observed causal structure represented by a chain graph. 

% This paper is organized as follows. Section 2 briefly introduces related works, and Section 3 proposes our VAE model, including its derivation and the proposed metric. Section 4 shows the results of numerical simulations. Concluding remarks follow in Section 5.

\textbf{Notation.} Let $\bv = (\bv_1, \bv_2, \cdots, \bv_p) \in \mbR^p$ be a $p$-dimensional vector and $\bfa = (\bfa_1, \cdots, \bfa_m)$ be a tuple where $\bfa_j \in \{1, \cdots, p\}$. $\bv_\bfa \coloneqq (\bv_{\bfa_1}, \bv_{\bfa_2}, \cdots, \bv_{\bfa_m})$ denotes a subvector of $\bv$ sliced by $\bfa$. Let $\sigma(1), \sigma(2), \cdots, \sigma(K)$ form a partition of $\{1,2,\cdots,p\}$ and assume that each $\sigma(j)$ is the ordered tuple whose elements are increasing. For a set $I = \{i_1, i_2, \cdots, i_k\} \subseteq \{1, \cdots, K\}$, $\sigma(I) \coloneqq \sigma(i_1) \oplus \cdots \oplus \sigma(i_k)$, where $\oplus$ is concatenation of tuples and $i_1 < i_2 < \cdots < i_k$. In particular, for $I=\{1,\cdots, K \}$ we denote $\sigma(I)$ simply by $\sigma$ and we call $\sigma$ the permutation inducing $K$-block partion on $\mathbb{R}^p$.
% $\bv_{\sigma} := \bv_{\sigma(\{1, 2, \cdots, K\})} = (\bv_{\sigma(1)}, \bv_{\sigma(2)}, \cdots, \bv_{\sigma(K)})$. 

% \textbf{Notation.} Let $\bv = (\bv_1, \bv_2, \cdots, \bv_p) \in \mbR^p$ be a $p$-dimensional vector and $\bfa = \{\bfa_1, \cdots, \bfa_m\}$ be a set of indices where $\bfa \subseteq \{1, \cdots, p\}$. $\vert \bfa \vert$ indicates the cardinality of $\bfa$ and $\bv_\bfa \coloneqq (\bv_{\bfa_1}, \bv_{\bfa_2}, \cdots, \bv_{\bfa_m})$. Let $\{\sigma(1), \cdots, \sigma(K)\}$ form a partition of $\{1,2,\cdots,p\}$ and additionally, we assume components of the partition are sorted in the increasing order without loss of generality. Then, we define $\bv_{\sigma}$ as a $K$-block partition of $\bv$ where $\bv_{\sigma} = (\bv_{\sigma(1)}, \bv_{\sigma(2)}, \cdots, \bv_{\sigma(K)})$. Lastly, for a set $I \subseteq \{1, 2, \cdots, K\}$, $\sigma(I) \coloneqq \cup_{l \in I} \sigma(l)$.

% be a set where its elements are indices of $v$. If $\sigma(i) \cap \sigma(j) = \varnothing$ for $i, j \in \{1, \cdots, K\}$ and $\sum_{j=1}^K \vert \sigma(j) \vert = p$, then we define $v_{\sigma}$ as a $K$-block partition of $v$ and $v_{\sigma} = (v_{\sigma(1)}, v_{\sigma(2)}, \cdots, v_{\sigma(K)})$. 

\begin{figure*}[h]
    \centering
    \subfigure[GM1]{\resizebox{0.3\textwidth}{!}{
    \begin{tikzpicture}[shorten >=1pt,auto,thick,main node/.style={circle,fill=blue!20,draw,font=\sffamily\Large\bfseries}]
    % nodes 
    \node[state] (x) {$\bx = D(\bg,\bu)$};
    \node[state] (g) [left = of x] {$\bg$};
    \node[state] (u) [right = of x] {$\bu$};
    % Directed edge
    \path (g) edge (x);
    \path (u) edge (x);
    \end{tikzpicture}
    }}
    \hfill
    \subfigure[GM2]{\resizebox{0.3\textwidth}{!}{
    \begin{tikzpicture}[shorten >=1pt,auto,thick,main node/.style={circle,fill=blue!20,draw,font=\sffamily\Large\bfseries}]
    % nodes 
    \node[state] (u) {$\bu$};
    \node[state] (g) [right = of u] {$\bg$};
    \node[state] (x) [right = of g] {$\bx = D(\bg)$};
    % Directed edge
    \path (u) edge (g);
    \path (g) edge (x);
    \end{tikzpicture}
    }}
    \hfill
    \subfigure[GM3]{\resizebox{0.3\textwidth}{!}{
    \begin{tikzpicture}[shorten >=1pt,auto,thick,main node/.style={circle,fill=blue!20,draw,font=\sffamily\Large\bfseries}]
    % nodes 
    \node[state] (g) {$\bg$};
    \node[state] (x) [right = of g] {$\bx = D(\bg)$};
    \node[state] (u) [right = of x] {$\bu$};
    % Directed edge
    \path (g) edge (x);
    \path (x) edge (u);
    \end{tikzpicture}
    }}
    \caption{Various generative model (GM) structures.}
    \label{fig:generative}
\end{figure*}
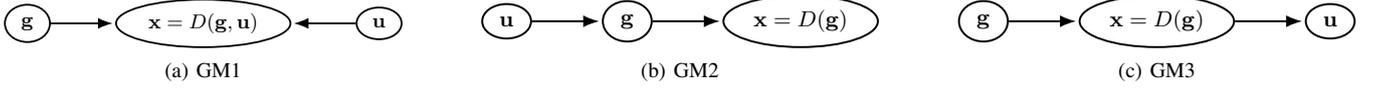

\section{Assumptions and Model Structure}
\label{sec:2}

\subsection{Data Generating Process}
Let $\bx = (\bx_1, \cdots, \bx_p) \in \mathcal{X} \subset \mathbb{R}^p$, $\bg = (\bg_1, \cdots, \bg_d) \in \mathbb{R}^d$, $\bu = (\bu_1, \cdots, \bu_d) \in [0, 1]^d$ with $d < p$ be the observation, the ground-truth factor generating $\bx$, and the annotation vector of $\bx$. The generation and causal structure in $\bx$ entirely depends on $\bg$. It is assumed that $\bg$ is unobservable, and the causality of $\bg$ is identified only through $\bu$. 
% Note that they have the same number of components of the partition. 

% \begin{figure}[ht]
%     \centering
%     \subfigure[]{
%     \includegraphics[width=0.375\columnwidth]{fig/chaingraph1.png} 
%     \label{fig:GTDAG}
%     }
%     \hspace{-3mm}
%     \subfigure[]{
%     \includegraphics[width=0.57\columnwidth]{fig/chaingraph2.png} 
%     \label{fig:cdg_dag}
%     }
%     \caption{Examples of graph structures when $K=3$. Dashed edges indicate induced edges by causations of $\bg_\pi$, and deterministic nodes are denoted as a dotted line. (a) The ground-truth data-generating process and the chain graph-structured $\bx_\sigma$. (b) The data-generating process of the generative model. Red edges represent CDG (see Section \ref{sec:2.3}).}
%     \label{fig:DAG}
% \end{figure}

Figure \ref{fig:generative} shows three popular generative models (GM1, GM2, GM3) for causal disentanglement learning. In GM1 \cite{Kingma2014SemisupervisedLW, Zheng2018DisentanglingLS, Hajimiri2021SemiSupervisedDO}, $\bg_i, i=1,\cdots,d$ are mutually independent since the annotation vector $\bu$ is directly used to generate the observation $\bx$. In GM2 \cite{yang2021causalvae}, $\bu$ determines the conditional distribution of the ground-truth factor. The proposed model follows GM3 employed by \cite{suter2019robustly, shen2022weakly, reddy2022causally}. GM3 differs from GM2 in that $\bu$ is not explicitly used to construct the latent space. Rather, $\bu$ is a recognized property of $\bx$, the annotation. \cite{Locatello2020Disentangling} proposes the supervised regularization ensuring that the learned latent representation is disentangled. 

% First, we assume that all $\bx \in \mathcal{X}$ follow a distributional-free data-generating process of Figure \ref{fig:GTDAG} ($\bg_{\pi(j)} \rightarrow \bx_{\sigma(j)}$, red edges of Figure \ref{fig:GTDAG}), which represents the blockwise correspondence from $\bg_\pi$ to $\bx_\sigma$. \cite{lachapelle2022disentanglement} shows that this assumption is required to infer the relationship among objects in the image. Note that what is modeled in the data-generating process of Figure \ref{fig:GTDAG} is only causal structures of the ground-truth factors, observations, and labels. 

\begin{assumption}[Blocked representation] \label{assump:block}
Partition tuples $\sigma(j), \pi(j), j=1,\cdots,K$ are given such that there are (block) causal relationships, $\bg_{\pi(j)} \rightarrow \bx_{\sigma(j)}, j=1,\cdots,K$.
\end{assumption}
\cite{lachapelle2022disentanglement} shows that Assumption \ref{assump:block} is required to infer the relationship among objects in the image. We denote the permutations inducing the $K$-block partition for $\bx$ and $\bg$ by $\sigma$ and $\pi$ and let $\sigma$ and $\pi$ be $K$-block consecutive partitions without loss of generality. 
% Also, the rearranged vector of $\bx$ and $\bg$ with $\sigma$ and $\pi$ are denoted as $\bx_\sigma$ and $\bg_\pi$.

% Note that $\bg$ and $\mbE[\bu | \bx]$ have the same causal structure by \eqref{eq:supervision}. The conditional independence $\bu \indep \bg | \bx$ holds from GM3 of Figure \ref{fig:generative}, and is if and only if that $\mbE[m_1(\bu) \cdot m_2(\bu) | \bx] = \mbE[m_1(\bu) | \bx] \cdot \mbE[m_2(\bu) | \bx]$ for all measurable functions $m_1$ and $m_2$. Since \eqref{eq:supervision} implies that $\bg$ is $\cF(\bx)$-measurable, the above statement is always true. 

\subsection{Embedding Causality} 
\label{sec:2.1}

We introduce the Markovian SCM of $\bg$ \cite{pearl2009causality, peters2017elements}, denoted by $\cM(\bg, {\bf e}, {\bf F}, P_{\bf e})$, where $\bg$ and ${\bf e} \in \mathbb{R}^d$ are the endogenous and exogenous variables, ${\bf F}$ is a set of structural equations, and $P_{\bf e}$ is a probability measure of ${\bf e}$. 
\begin{assumption} \label{assump:SCM}~\\
1. [Identifiable SCM] $\cM(\bg, {\bf e}, {\bf F}, P_{\bf e})$ is identifiable with the directed acyclic graph (DAG) $\mathcal{G}$ derived from ${\bf F}$. \\
2. [Unconfoundedness] The observation $\bx$ is generated based on $\cM(\bg, {\bf e}, {\bf F}, P_{\bf e})$ and $\bx_{\sigma(j)}$ depends on only ${\bf{g}}_{\pi(j)}$, for $j=1,\cdots,K$. 
\end{assumption}
% It is assumed that $\cM(\bg, {\bf e}, {\bf F}, P_{\bf e})$ is identifiable with the directed acyclic graph (DAG) $\mathcal{G}$ derived from ${\bf F}$. Note that the observation $\bx$ is generated based on $\cM(\bg, {\bf e}, {\bf F}, P_{\bf e})$ and $\bx_{\sigma(j)}$ depends on only ${\bf{g}}_{\pi(j)}$, for $j=1,\cdots,K$. 

It is ideal to employ $\cM(\bg, {\bf e}, {\bf F}, P_{\bf e})$ as latent variables of CDG-VAE. However, we reduce $\cM(\bg, {\bf e}, {\bf F}, P_{\bf e})$ to a simple non-linear structure equation model because $P_{\bf e}$ is unknown and ${\bf F}$ leads to heavy computational expense in training our generative model. Let $B \in \{0, 1\}^{d \times d}$ be a binary adjacency matrix whose element indicates the existence of directed edges of $\mathcal{G}$ (i.e., the causal relationships between $K$ block subvectors of $\bg_\pi$). For a subvector with an index $s$, the set of its non-descendants is denoted as $ND(s)$, and the set of its descendants is denoted as $Des(s)$. 

The reduced non-linear structural equation model \cite{yu2019dag, yang2021causalvae, shen2022weakly} is 
\bea \label{eq:scm}
f^{-1} (\bz) = B^\top f^{-1} (\bz) + \bepsilon,
%\\ \bz &=& f((I - B^\top)^{-1} \bepsilon) =: F(\bepsilon; f, B), \nonumber,
\eea
where $\bz \in \mathbb{R}^d$ is an induced latent variable by an element-wise invertible function $f$ and the $d$-dimensional standard normal distribution random variable $\bepsilon$. The distribution of latent variables in CDG-VAE is defined based on the entailed distribution of \ref{eq:scm} and the distribution of $\bepsilon$, $p(\bepsilon)$. The modeling of $p(\bz)$ through $B$ is distinguished from conventional VAE models whose prior distributions are given without considering causality. Here, $B$ is treated as a known matrix.  

\subsection{Derivation of CDG-VAE} 

The decoder of CDG-VAE is given by $p(\bx_\sigma|\bz;\theta,\beta) = \mathcal{N}(\bx_\sigma | D(\bz;\theta)_\sigma, \beta \cdot I)$, where $D: \mbR^d \mapsto \mbR^p$ is a function parameterized with $\theta$, $I$ is the $p \times p$ identity matrix, and $\beta > 0$ is the non-trainable observation noise. The rearranged vector of $D(\bz; \theta)$ with $\sigma$ is denoted as $D(\bz; \theta)_\sigma$, where $D(\bz; \theta)_\sigma = (D(\bz; \theta)_{\sigma(1)},\cdots,D(\bz; \theta)_{\sigma(K)})$.

% As an extension, the latent space can be modeled $d+p$ dimension, where additional $p$ coordinates represent features of observations that are unexplained by a causal relationship.

The proposal distribution $q(\bepsilon|\bx;\phi)$ of CDG-VAE is given by $\mathcal{N}\big(\bepsilon | \mu(\bx;\phi), diag(\sigma^2(\bx;\phi))\big)$, where $\mu:\mbR^p \mapsto \mbR^d$, $\sigma^2:\mbR^p \mapsto \mbR_+^d$ are neural networks parameterized with $\phi$, and $diag(a), a \in \mbR^d$ denotes a diagonal matrix with diagonal elements $a$. Based on the proposal distribution, the negative ELBO (Evidence of Lower BOund) is written as
\bea \label{eq:elbo}
\mathcal{L}(\bx; \theta, \phi, f) &\coloneqq& \mbE_{q} \left[\frac{1}{2} \parallel \bx_\sigma - D(F(\bepsilon; f, B);\theta)_\sigma \parallel^2\right] \nonumber \\
&+& \beta \cdot \mathcal{KL}(q(\bepsilon|\bx; \phi) \| p(\bepsilon))
\eea
where $q$ is the proposal distribution, $F(\bepsilon; f, B) := f((I - B^\top)^{-1} \bepsilon)$, $f$ is parameterized with the flow-based model \cite{rezende2015variational, behrmann2019invertible}, $\mathcal{KL}(q\|p)$ denotes Kullback-Leibler divergence from $p$ to $q$, and constant terms are omitted (see Appendix A.1 for detailed derivation). See Appendix A.1 for detailed covariance structures of $\bz$. The rearranged vector of $\bz$ with $\pi$ is denoted as $\bz_\pi$. 

\begin{figure}[h]
    \centering
    \includegraphics[width=0.7\columnwidth]{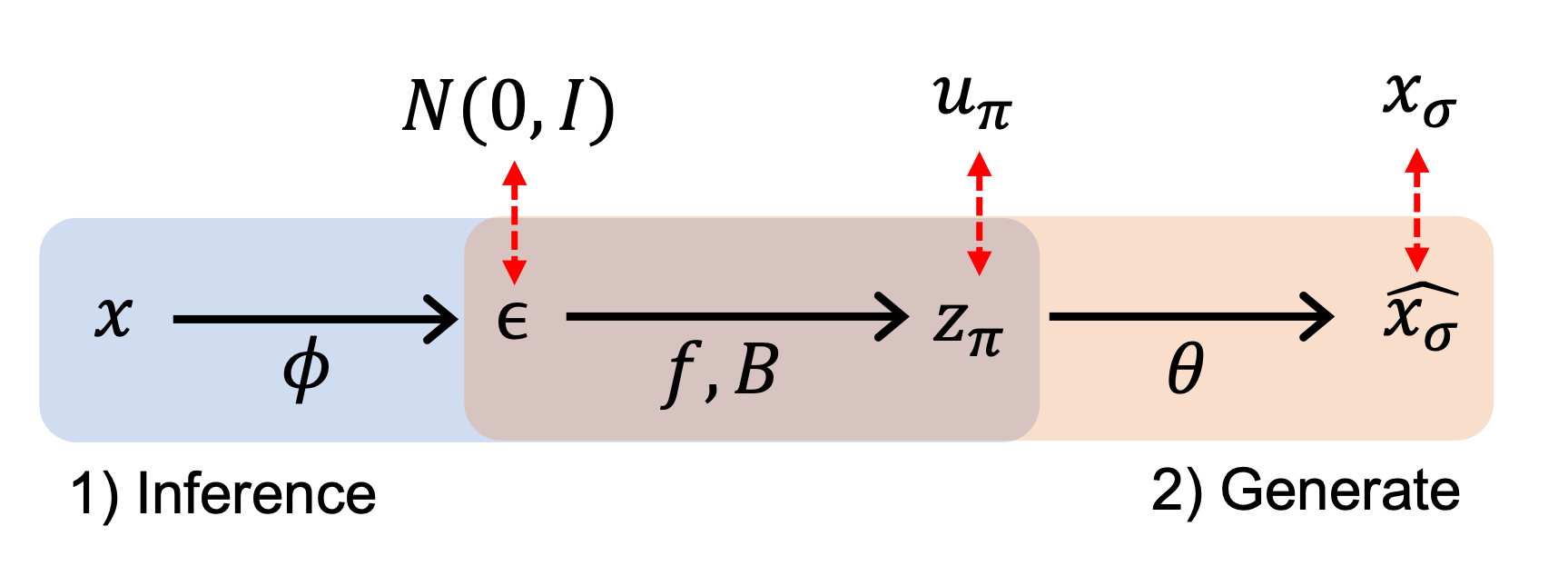}
    \caption{Model structure of CDG-VAE.}
    \label{fig:diagram}
\end{figure}

\subsection{Reguarization for Disentangled Representation}
\label{sec:2.2}

In the supervision setting \cite{Locatello2020Disentangling} of GM3, we adopt Assumption \ref{assump:supervision}, which was adopted in semi-supervised causal disentanglement learning \cite{shen2022weakly}. It implies the generative model with the type of GM3, the conditional independence $\bu \indep \bg | \bx$, because $\bg$ is $\cF(\bx)$-measurable. 

\begin{assumption}[Informative supervision] \label{assump:supervision}
$\bg_{i} = \mbE[\bu_{i} | \bx]$, $\forall i$.
% for $i=1,\cdots,d$. 
\end{assumption}
% In the supervision setting \cite{Locatello2020Disentangling} of GM3, we assume that
% \bea \label{eq:supervision}
% \bg_{i} = \mbE[\bu_{i} | \bx],
% % \bg_{i} = \mbE[\bu_{i} | \bx_{\sigma(j)}] = \mbE[\bu_{i} | \bx],
% \eea
% for $i=1,\cdots,d$. 

% In general, the dimension size of $\bx$ is larger than that of $\bg$, it is not an unrealistic assumption that $\bg$ is $\cF(\bx)$-measurable. 

Since the ground-truth factors are causally related, the representation of our encoder is entangled as the ground-truth factors \cite{trauble2021disentangled}. With $F(\mu(\bx;\phi); f, B)$, we aim to train the encoder parameters $(\phi, f)$ to approximate the SCM of $\mbE[\bu|\bx]$ by \eqref{eq:scm}. The following definition formalizes this concept.

\begin{definition}[Disentangled Representation \cite{shen2022weakly}] \label{def:disentangle}
For $i = 1, \cdots, d$, suppose that there exists a 1-to-1 function $h_i$ such that
\bean
F(\mu(\bx;\phi); f, B)_i = h_i^{-1}(\mbE[\bu_i|\bx]),
\eean
where $(\phi, f)$ are parameters of an encoder of a generative model, and the subscript $i$ denotes the $i$th element. Then the generative model is said to have a disentangled representation with respect to $\mbE[\bu|\bx]$.
\end{definition}

Note that Definition \ref{def:disentangle} implies that the disentangled representation only depends on the encoder of VAE. We adopt the supervised loss to obtain the disentangled representation, which aligns the representation and annotation vector \cite{Locatello2020Disentangling}. Our final objective is minimizing
\bea \label{eq:obj}
% \min_{\theta, \phi, f} 
\mbE_{\bx} [\mathcal{L}(\bx; \theta, \phi, f)] + \lambda \cdot \mbE_{\bx, \bu} [\ell(F(\mu(\bx;\phi); f, B), \bu)]
\eea
with respect to $(\theta, \phi, f)$, where $\lambda > 0$ is the tuning parameter and $\mbE_{\bx}$, $\mbE_{\bx, \bu}$ indicate expectations for the marginal and joint distribution of the dataset, respectively. Note that \eqref{eq:obj} is easily extended to the semi-supervised learning model because the negative ELBO $\cL$ and the supervised loss $\ell$ are decoupled. In this paper, $\ell$ is the cross-entropy loss with the sigmoid function $h_i$ for $\bu \in [0, 1]^d$ \cite{shen2022weakly}. 
% For detailed choices of $\ell$ and $h_i$, please refer \cite{shen2022weakly}.

% Note that our objective \eqref{eq:obj} is easily extended to the semi-supervised setting because negative ELBO and supervised loss are decoupled. Here, the supervised loss $\ell$ can be defined as L2 loss $\parallel F(\mu(\bx;\phi); f, B) - \bu \parallel^2$ for continuous label $\bu$, or cross-entropy $- \sum_{i=1}^d \bu_i \cdot \log \sigma(F(\mu(\bx;\phi); f, B)_i) + (1 - \bu_i) \cdot \log (1 - \sigma(F(\mu(\bx;\phi); f, B)_i))$ for $[0, 1]$ bounded or binary label where $\sigma(\cdot)$ is sigmoid function \cite{shen2022weakly}. And $g_i$ is the identity function for continuous labels and the sigmoid function for bounded or binary labels. 

\section{Properties of CDG-VAE}
\label{sec:2.3}

\subsection{Causally Disentangled Generation}

% Before we start, we need to define the notation for hierarchical dependencies. For $j=1,\cdots,p$, $\bx_{\sigma(ND(j))}$ denotes non-descendant and $\bx_{\sigma(Des(j))}$ denotes descendant chain components of $\bx_{\sigma(s)}$, where $j \in \sigma(s), s=1,\cdots,K$. And the non-descendant and descendant chain components are defined analogously for $\bz_\pi$ and $D(\bx;\theta)_\sigma$.

The disentanglement in the latent space has been studied with the link of explainability of VAE \cite{higgins2016beta, bouchacourt2018multi, dupont2018learning, locatello2019disentangling}. The generative power of the latent space is theoretically investigated by the dimension of an activated latent space \cite{An2023CustomizedLS}, which indicates a regular condition of the encoder. However, the disentangled latent space obtained by the encoder does not directly guarantee the generation of causally plausible data. We discover that the disentangled generation in the decoder is necessary.

A simple example shows that $D(\bz;\theta)_{\sigma(ND(K))}$ determined by $\bz_{\pi(ND(K))}$ can be affected by $do$-intervention on $\bz_{i}, i \in \pi(K)$ due to entangled decoder structure (see examples in Section \ref{sec:4.2} and Appendix A.7). However, in the disentangled latent space, a total causal effect does not exist from $\bz_{i}, i \in \pi(K)$ to $\bz_{j}, j \in \pi(ND(K))$ because there is no directed path \cite{Peters2017ElementsOC}. This result implies that causal generation requires two types of disentanglement in encoding and decoding processes, respectively. We propose a new definition of the causally disentangled generation.

% Specifically, suppose the first layer of the decoder mean function is modeled with MLP. In that case, the disentangled representation information is delivered to the decoder output in a mixed way because of the dense weight matrices of MLP.

\begin{definition}[Causally Disentangled Generation (CDG)] \label{def:CDG} 
% Suppose that a generative model has a causally disentangled representation. 
Suppose a generative model has a disentangled representation (Definition \ref{def:disentangle}). Let $D(\cdot;\theta): \mbR^d \mapsto \mbR^p$ be a decoder mean vector of the model where the input is denoted as $\bz \in \mbR^d$. Then the model is causally disentangled generative if, for $s=1,\cdots,K$, $D(\bz; \theta)_{\sigma(s)}$ is independent to $\bz_{\pi(l)}, l \neq s$, given $\bz_{\pi(s)}$.
% $D(\bz; \theta)_{\sigma(s)}$ is only determined by $\bz_{\pi(s)}$ and is independent to $\bz_{\pi(l)}, l \neq s$, under no intervention.
\end{definition}

% \begin{figure}[ht]
%     \centering
%     \resizebox{0.6\columnwidth}{!}{
%     \begin{tikzpicture}[shorten >=1pt,auto,thick,main node/.style={circle,fill=blue!20,draw,font=\sffamily\Large\bfseries}]
%     \node[state] (z) {$\bz_{\pi(j)}$};
%     \node[state] (x) [right = of z] {$D(\bz;\theta)_{\sigma(j)}$};

%     \path (z) edge (x);
%     \end{tikzpicture}
%     }
%     \caption{The CDG process for $j=1,\cdots,K$.}
%     \label{fig:cdg_dag}
% \end{figure}

% Since the encoder is required to make the latent variable $\bz$ to be causally disentangled (D1) and the decoder needs to keep disentangled information separated ((D1), (D2), (D3)).
% Definition \ref{def:CDG} implies that causal disentanglement learning depends on both the encoder and the decoder. 

Definition \ref{def:CDG} implies that CDG mimics the causal relationship of $\bg_{\pi(j)} \rightarrow \bx_{\sigma(j)}$ as $\bz_{\pi(j)} \rightarrow D(\bz;\theta)_{\sigma(j)}$ for $j=1,\cdots,K$. And the following Proposition \ref{prop:decoder} demonstrates the sufficient condition of the decoder function class satisfying CDG. 
% If CDG is satisfied, then the $j$th decoder partition should depend only on $\bz_{\pi(j)}$. 
% Definition \ref{def:CDG} implies that CDG mimics the data-generating process of Figure \ref{fig:GTDAG} as Figure \ref{fig:cdg_dag} ($\bz_{\pi(j)} \rightarrow D(\bx;\theta)_{\sigma(j)}$, red edges in Figure \ref{fig:cdg_dag}). 

\begin{proposition}[Sufficient Condition for CDG] \label{prop:decoder}
Suppose a generative model has a disentangled representation (Definition \ref{def:disentangle}). Let $D(\cdot;\theta): \mbR^d \mapsto \mbR^p$ be a decoder mean vector of the model. If the decoder structure of the model satisfies $D(\bz;\theta)_\sigma := \Big(D(\bz_{\pi(1)};\theta_1), \cdots, D(\bz_{\pi(1)};\theta_K)\Big)$, where $\theta=(\theta_1,\cdots,\theta_K)$, and $D(\cdot;\theta_j): \mbR^{|\pi(j)|} \mapsto \mbR^{|\sigma(j)|}$ is a function parameterized with $\theta_j$ for $j=1,\cdots,K$, then the model satisfies Definition \ref{def:CDG}.
% (2) $D(\bz;\theta)_\sigma := \Big(D(\bz;\theta_1), \cdots, D(\bz;\theta_K)\Big)$, where $\theta=(\theta_1,\cdots,\theta_K)$, and $D(\cdot;\theta_j): \mbR^d \mapsto \mbR^{|\sigma(j)|}, j=1,\cdots,K$ is a neural network parameterized with $\theta_j$ where the first layer is MLP, and the first layer weight matrix columns of which indices correspond to $\pi(l), l \neq j$ consist of all zeros.
% If the decoder of the model satisfies either (1) or (2), then the model satisfies Definition \ref{def:CDG}.
\end{proposition}

\begin{proof}
If a generative model satisfies Proposition \ref{prop:decoder}, the $j$th decoder partition $D(\bz_{\pi(j)};\theta_j)$ (which corresponds to $D(\bz;\theta)_{\sigma(j)}$) takes only a block-partitioned $\bz_{\pi(j)}$ as input, for $j=1,\cdots,K$. Therefore, for $j=1,\cdots,K$, since $\bz_{\pi(j)}$ is the direct cause of $D(\bz_{\pi(j)};\theta_j)$, $D(\bz_{\pi(j)};\theta_j)$ is independent to $\bz_{\pi(i)}, i \neq j$, given $\bz_{\pi(j)}$.
\end{proof}

% Proposition \ref{prop:decoder} can be rewritten under a general additive model. Instead, Proposition \ref{prop:decoder} focuses on the disentanglement of the decoder in neural network structural modeling. For general decoder functions, we can remove the non-direct causal effects of the latent variables by introducing the group LASSO penalty.

\begin{assumption}[Faithfulness] \label{assump:faithfulness}
The entailed distribution of \eqref{eq:scm} is faithful with respect to the graph induced by $B$.
\end{assumption}

\begin{proposition}[Existence of Total Causal Effect (TCE)] \label{prop:tce}
Suppose a generative model has a disentangled representation (Definition \ref{def:disentangle}) and satisfies CDG (Definition \ref{def:CDG}). Let $D(\cdot;\theta): \mbR^d \mapsto \mbR^p$ be a decoder mean vector of the model where the input is denoted as $\bz \in \mbR^d$. For $s=1,\cdots,K$, under Assumption \ref{assump:faithfulness},

1. there is no total causal effect from $\bz_i$ to $D(\bz; \theta)_j$, for all $i \in \pi(s)$ and $j \in \sigma(ND(s))$.
% (\textbf{Interventional robustness}). 

2. if there is a directed path from $\bz_i$ to $\bz_j$ for some $i \in \pi(s)$ and $j \in \pi(l)$ where $l \in \{s\} \cup Des(s)$, then there is a total causal effect from $\bz_i$ to $D(\bz; \theta)_k$, for all $k \in \sigma(l)$.
% $j \in \sigma(s) \oplus \sigma(Des(s))$. 
% (\textbf{Counterfactual generativeness}).
\end{proposition}

Proposition \ref{prop:tce} states that the existence of TCE can be investigated by intervening on the latent variable if the causal structure identified by known $B$ is embedded precisely in the latent space and a model satisfies CDG. Since TCE determines the positiveness of the causal effect, Proposition \ref{prop:tce} motivates us to check the validity of CDG by measuring the causal effect (see Section \ref{sec:2.4}). Due to the computational issue, we estimate the causal effect on the annotation vector instead of block partitions, based on the causal relationship from $D(\bz;\theta)_\sigma$ to $\bu_\pi$.
% ($D(\bz; \theta)_{\sigma(j)} \rightarrow \bu_{\pi(j)}$, for $j=1,\cdots,K$). 

% Further, we assume the causal relationship $\bx_{\sigma(j)} \rightarrow \bu_{\pi(j)}$ and $D(\bz; \theta)_{\sigma(j)} \rightarrow \bu_{\pi(j)}$, for $j=1,\cdots,K$. By introducing this additional causal relationship, we can measure the degree of causally disentangled generativeness of a generative model by our proposed metric (see Section \ref{sec:2.4}). It implies that $\mbE[\bu_{i} | D(\bz; \theta)_{\sigma(j)}] = \mbE[\bu_{i} | D(\bz; \theta)]$, for ${i} \in \pi(j)$, $j=1,\cdots,K$.

First, we define the average causal effect of the latent variable on the annotation vector (the counterfactual quantity corresponding to the ground-truth factors \cite{reddy2022causally}) and then propose necessary conditions for CDG based on the average causal effect. We denote $z^{(1)}, z^{(2)}$ as the vector of maximum and minimum values of latent variables given the observed dataset, respectively, and denote $\pi(s)_{-i}$ as the partition tuple $\pi(s)$ without an index $i$.
% observed maximum and minimum values of the intervened node $\bz_{\pi(s)}, s=1,\cdots,K$ are given as $z_{\pi(s)}^{(max)}, z_{\pi(s)}^{(min)}$, respectively.

\begin{definition}[Average Causal Effect (ACE)] \label{def:ace}
Suppose that $\bz_i, i \in \pi(s), s=1,\cdots,K$ is intervened with $z^{(1)}_i$ and $z^{(2)}_i$. Then, for $c=1,\cdots,d$, the average causal effect of $\bz_i$ on the annotation vector $\bu_c$ given $z_{\pi(ND(s)) \oplus \pi(s)_{-i}}$ is defined as
\bean 
&& ACE(\bu_c, \bz_i, \bz_{\pi(ND(s)) \oplus \pi(s)_{-i}} = z_{\pi(ND(s)) \oplus \pi(s)_{-i}}) \\
&\coloneqq& \Big\vert \mbE[\bu_c \vert z_{\pi(ND(s)) \oplus \pi(s)_{-i}}, do(\bz_i \coloneqq z^{(1)}_i)] \\
&-& \mbE[\bu_c \vert z_{\pi(ND(s)) \oplus \pi(s)_{-i}}, do(\bz_i \coloneqq z^{(2)}_i)] \Big\vert.
\eean
\end{definition}
% Instead of the absolute distance, any other distance function can be applied. 

\begin{proposition}[Necessary Conditions for CDG] \label{prop:monotonic}
For $i \in \pi(s), s=1,\cdots,K$, assume that arbitrary $x$ and $z_{\pi(ND(s)) \oplus \pi(s)_{-i}}$ are given. $z^{(j)}_{(i,z_{\pi(ND(s))},x)}$ denotes $\bz$ defined in \eqref{eq:scm} under intervention $do(\bz_i \coloneqq z^{(j)}_i)$ given $x$ and $z_{\pi(ND(s)) \oplus \pi(s)_{-i}}$, for $j=1, 2$. Suppose a generative model has a disentangled representation (Definition \ref{def:disentangle}). For $c = 1, \cdots, d$, under Assumption \ref{assump:faithfulness}, if the model satisfies CDG (Definition \ref{def:CDG}) and

1. $c \in \pi(ND(s))$, then
\bean
% ACE(c, s, \bz_{\pi(ND(s))} = z_{\pi(ND(s))}) = 0.
ACE(\bu_c, \bz_i, z_{\pi(ND(s)) \oplus \pi(s)_{-i}}) = 0.
% \mbE[\bu_c \vert z_{\pi(ND(s))}, do(z_{\pi(s)}^{(1)})] = \mbE[\bu_c \vert z_{\pi(ND(s))}, do(z_{\pi(s)}^{(2)})].
\eean

2. there is a directed path from $\bz_i$ to $\bz_c$ where $c \in \pi(l)$ and $l \in \{s\} \cup Des(s)$, then
\bean
% 0 &\leq& ACE(c, s, \bz_{\pi(ND(s))} = z_{\pi(ND(s))}) \\
% &\leq& \mbE_{p(\bx)} \Bigg[ \Big\vert \mbE[\bu_c \vert z^{(max)}] - \mbE[\bu_c \vert z^{(min)}] \Big\vert \Bigg] \neq 0.
0 &<& ACE(\bu_c, \bz_i, z_{\pi(ND(s)) \oplus \pi(s)_{-i}}) \\
&\leq& \mbE_{p(\bx)} \Big\vert \mbE[\bu_c \vert z^{(1)}_{(i,z_{\pi(ND(s))},\bx)}] - \mbE[\bu_c \vert z^{(2)}_{(i,z_{\pi(ND(s))},\bx)}] \Big\vert,
% \mbE[\bu_c \vert z_{\pi(ND(s))}, do(z_{\pi(s)}^{(1)})] \neq \mbE[\bu_c \vert z_{\pi(ND(s))}, do(z_{\pi(s)}^{(2)})].
\eean
where $p(\bx)$ is the probability density function of $\bx$.
\end{proposition}
% \begin{proof}
% See Appendix XXX.
% \end{proof}

% Lastly, the construction of a decoder satisfying Proposition \ref{prop:decoder} is easy because an arbitrary decoder of which the first layer is MLP can be modified to satisfy Proposition \ref{prop:mlp} (see Appendix \ref{app:2}). We discover that all latent dimensions are always active under Proposition \ref{prop:decoder}.
% as the CDG process of Figure \ref{fig:cdg_dag} imposes the blockwise correspondence from $\bz_\pi$ to $D(\bz;\theta)_\sigma$,

% Note that we only restrict the function class of the decoder mean vector. We do not surgery the ELBO or adopt any additional regularization such as mutual information. And we call our proposed VAE model `CDG-VAE'.

\subsection{Causal Disentanglement Metric}
\label{sec:2.4}

Based on Proposition \ref{prop:monotonic}, we propose a metric that can evaluate how much a model is causally disentangled generative.

\begin{definition}[Causal Disentanglement Metric (CDM)] \label{def:cdm}
For $c=1,\cdots,d$ and $i \in \pi(s), s=1,\cdots,K$, the causal disentanglement metric (CDM) is defined as 
\bean
% CDM(c, s) = \mbE_{\bz_{\pi(ND(s))}} [ACE(c, s, \bz_{\pi(ND(s))})] \geq 0.
CDM(c, i) \coloneqq \mbE [ACE(\bu_c, \bz_i, \bz_{\pi(ND(s)) \oplus \pi(s)_{-i}})],
% CDM(c, s) &=& \mbE_{\bz_{\pi(ND(s))}} [ACE(c, s, \bz_{\pi(ND(s))})] \\
% &=& \int ACE(c, s, \bz_{\pi(ND(s))} = z_{ND(s)}) \cdot p(z_{ND(s)}) dz_{ND(s)} 
\eean
where $\mbE$ indicates the expectation with respect to $\bz_{\pi(ND(s)) \oplus \pi(s)_{-i}}$.
\end{definition}
% The definition of CDM implies the followings. 

First, CDM for the first case of Proposition \ref{prop:monotonic} measures interventional robustness \cite{suter2019robustly}. If a generative model satisfies CDG, ACE should be zero by Proposition \ref{prop:monotonic} (i.e., no total causal effect), and CDM is also exactly zero. Thus, $CDM(c, i) > 0$ implies that the model is not interventional robust. 

Next, CDM for the second case of Proposition \ref{prop:monotonic} measures counterfactual generativeness \cite{reddy2022causally}. Suppose a generative model satisfies CDG. Then, ACE should be non-zero (i.e., total causal effect exists) and is the lower bound of the coverage for the counterfactual quantity $\bu_c$ by Proposition \ref{prop:monotonic}. Therefore, $CDM(c, i) = 0$ or $CDM(c, i) \approx 0$ implies that the model lacks counterfactual generativeness.
% the data distribution of the $c$th chain component. 

% \cite{suter2019robustly} and \cite{reddy2022causally} considered disentangled causal mechanisms and proposed metrics for causal disentanglement. However, they assumed that there is no causal effect between latent variables. Also, the causal disentanglement metric of \cite{shen2022weakly} is not based on $do$-calculus with the SCM of ground-truth factors. 
\cite{suter2019robustly, reddy2022causally, shen2022weakly} have considered disentangled causal mechanisms and proposed metrics for causal disentanglement. Unlike our study, the existing studies do not deal with a causal effect between latent variables nor derive the metric based on the SCM of the ground-truth factors. In particular, our metric is theoretically justified by the necessary conditions for the CDG and the do-calculus of causal effects.
Therefore, CDM can be regarded as a generalized causal disentanglement metric under the arbitrary DAG structure of ground-truth factors. See Appendix A.4 for identification of CDM's upper and lower bounds. 

Also, the faithfulness condition (Assumption \ref{assump:faithfulness}) excludes the case where the directed path exists in $B$, but the total causal effect does not under. It implies that, under the faithfulness condition, our proposed metric CDM is valid to measure the causally disentangled generativeness of the model. However, without the faithfulness condition, we can only measure the interventional robustness of Proposition \ref{prop:monotonic}. Therefore, the faithfulness condition is required to measure the causal disentanglement of the model by the average causal effect.
% but also requires the ground-truth SCM.

\section{Related Work}
\label{sec:3}

\textbf{Active Latent Dimensions.} \cite{Burda2015ImportanceWA, Sicks2020AGL} propose the statistics to detect active latent dimensions, which encode useful information about the data and are significant in data generation. \cite{An2023CustomizedLS} shows that the posterior variance $\sigma^2(\bx;\phi)$ determines whether the latent dimension affects generated data. However, the definition of disentangled representation (Definition \ref{def:disentangle}) only depends on the deterministic component $\mu(\bx;\phi)$. Therefore, a latent dimension can be simultaneously non-active and disentangled. It is a critical issue because the causally plausible counterfactual samples can not be generated when non-active latent variables are intervened (see Appendix A.8 for an example). \cite{rezaabad2020learning} mitigates the non-activation issue by maximizing the additional mutual information regularization term. However, we discover that all latent dimensions are always active under Proposition \ref{prop:decoder}.
% In this sense, we argue that the disentangled representation does not guarantee that all latent dimensions are significant in data generation. 

\textbf{Supervised Causal Disentanglement Learning.} \cite{yang2021causalvae, shen2022weakly} propose causal disentanglement learning methods based on supervision setting \cite{Locatello2020Disentangling} and embed the SCM in the latent space to make the representation causally entangled. To obtain the disentangled representation, \cite{yang2021causalvae} aligns the latent variable and the annotation vector using KL-divergence and the customized prior, and \cite{shen2022weakly} regularizes the encoder. However, since \cite{yang2021causalvae, shen2022weakly} only constrain the encoder, their methods can not guarantee that generating causally plausible data is achievable.

% However, since \cite{yang2021causalvae, shen2022weakly} only constraint the encoder part, their methods can not guarantee that chain components $D(\bz;\theta)_{\sigma(ND(s))}$ are not affected by $do$-interventions on $\bz_{\pi(s)}$, even if a total causal effect does not exist from $\bz_{\pi(s)}$ to $\bz_{\pi(ND(s))}$. It implies that generating causally plausible data generation is unachievable with their methods.

\textbf{Causally-Aware Synthetic Data Generation.} Since tabular datasets are already well-structured, covariates (columns) are usually assumed to be causally related \cite{wen2021causal}. To exploit causations in the synthetic data generation, \cite{xu2019achieving, wen2021causal, van2021decaf} generate data in the order of causal topology, and consequently, they require the completely identified DAG, not the Partial DAG. However, from the observational data, the true causal graph of covariates can be identified only up to a Markov Equivalence Class (MEC), including undirected edges. Even though \cite{van2021decaf} proves that their generator converges to the right distribution for any graph belonging to MECs, incorrect edge directions have the potential risk of misunderstandings of causations. 
% \cite{xu2019modeling} proposed the synthetic data generation method using the mode-specific normalization method, and \cite{wen2021causal} extended the technique of \cite{xu2019modeling} by exploiting the causal structure between covariates. 
% And \cite{xu2019achieving, van2021decaf} focused on the fairness of synthetic data.

\section{Experiments}
\label{sec:4}

This section demonstrates that our model is causally disentangled in both the encoding and decoding processes. Our numerical experiments show that CDG-VAE can achieve two goals: 1) the causally plausible counterfactual generation under interventions on latent variables and 2) synthetic data generation preserving the observed causal structure. Furthermore, the performances of our model, sample efficiency, distributional robustness, and synthetic data quality, are presented with three downstream tasks. The code and appendix are available at \url{https://github.com/an-seunghwan/CDG-VAE}.

\subsection{Overview}
\label{sec:4.1}

\textbf{Dataset.} For evaluation, we consider two types of datasets, image and tabular. For an image dataset, a simulated pendulum dataset \cite{shen2022weakly, yang2021causalvae} is used. And \texttt{loan}, \texttt{adult}, and \texttt{covertype} datasets are used for real tabular datasets (see Appendix A.6 for detailed data descriptions). 

\noindent\textbf{Compared Models.} We train the vanilla VAE and InfoMax VAE \cite{rezaabad2020learning} based on the objective function \eqref{eq:obj} (see Appendix A.1 for detailed objective functions). We also compare existing disentangled generative models (CausalVAE \cite{yang2021causalvae}, DEAR \cite{shen2022weakly}) and synthesizers (TVAE \cite{xu2019modeling}, CTAB-GAN \cite{Zhao2021CTABGANET}).
% with their default settings except for latent dimension size. 

Note that all models are trained under the ground-truth causal graph, not a super-graph, because we numerically find that DEAR and CausalVAE are not able to discover ground-truth causal relationships. 
% Instead, learning coefficient values of the causal adjacency matrix. 
VAE, InfoMax VAE, and CDG-VAE share the same network architecture for the encoder; however, only CDG-VAE has the decoder structure of Proposition \ref{prop:decoder}. Notably, all models have the same size of the latent dimension. 

\subsection{Image Dataset}
\label{sec:4.2}

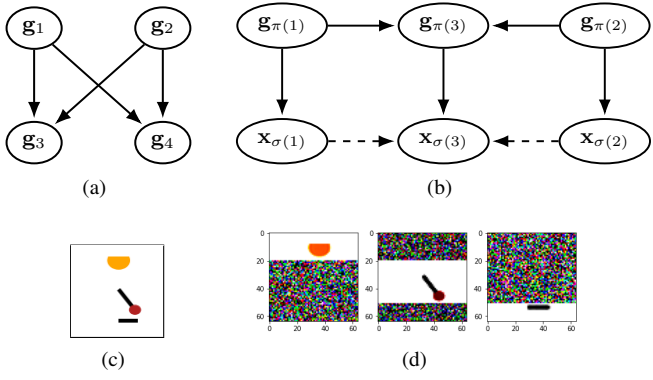
\begin{figure}[ht]
\centering
\subfigure[]{\resizebox{0.3\columnwidth}{!}{
\begin{tikzpicture}[shorten >=1pt,auto,thick,main node/.style={circle,fill=blue!20,draw,font=\sffamily\Large\bfseries}]
    % nodes 
    \node[state] (light) {$\bg_1$};
    \node[state] (angle) [right = of light] {$\bg_2$};
    \node[state] (length) [below = of light] {$\bg_3$};
    \node[state] (position) [below = of angle] {$\bg_4$};

    % Directed edge
    \path (light) edge (length);
    \path (light) edge (position);
    \path (angle) edge (length);
    \path (angle) edge (position);
\end{tikzpicture} \label{fig:pendulum_dag}}}
\hfill
\subfigure[]{\resizebox{0.64\columnwidth}{!}{
\begin{tikzpicture}[shorten >=1pt,auto,thick,main node/.style={circle,fill=blue!20,draw,font=\sffamily\Large\bfseries}]
    % nodes 
    \node[state] (g1) {$\bg_{\pi(1)}$};
    \node[state] (g3) [right = of g1] {$\bg_{\pi(3)}$};
    \node[state] (g2) [right = of g3] {$\bg_{\pi(2)}$};
    \node[state] (x1) [below = of g1] {$\bx_{\sigma(1)}$};
    \node[state] (x3) [below = of g3] {$\bx_{\sigma(3)}$};
    \node[state] (x2) [below = of g2] {$\bx_{\sigma(2)}$};

    % Directed edge
    \path (g1) edge (g3);
    \path (g2) edge (g3);
    \path (x1) [dashed] edge (x3);
    \path (x2) [dashed] edge (x3);
    
    \path (g1) edge (x1);
    \path (g2) edge (x2);
    \path (g3) edge (x3);
\end{tikzpicture}} \label{fig:chain_dag_pendulum}}
\subfigure[]{\frame{\includegraphics[width=0.14\columnwidth]{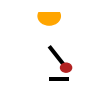}}}
\hspace{1cm}
\subfigure[]{\includegraphics[width=0.5\columnwidth]{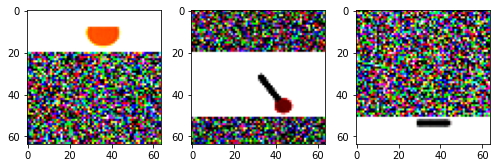}
\label{fig:sigma_pendulum}}
\caption{Pendulum dataset. (a) DAG of the ground-truth factors $\bg = (\bg_1, \cdots, \bg_4)$. (b) The causal relationships between $\bg_\pi$ and $\bx_\sigma$. Dashed edges indicate induced edges by causations of $\bg_\pi$. (c) An observation example. (d) From left to right, $\bx_{\sigma(1)}$ (the light), $\bx_{\sigma(2)}$ (the pendulum), and $\bx_{\sigma(3)}$ (the shadow).} 
\label{fig:chain_pendulum}
\end{figure}

In the pendulum dataset, there exist four ground-truth factors: $\bg_1$(light angle), $\bg_2$(pendulum angle), $\bg_3$(shadow length), and $\bg_4$(shadow position). These factors have the causal relationship given as the DAG structure of Figure \ref{fig:pendulum_dag}. Partition tuples of $\bg_\pi$ are $\pi(1) = (1), \pi(2) = (2)$, and $\pi(3) = (3, 4)$. Causal structures are visualized in Figure \ref{fig:chain_dag_pendulum}.

As in \cite{shen2022weakly}, we introduce random measurement noises in the generation of annotation vectors to make the pendulum dataset more realistic. Shadows of 20\% corrupted data are randomly generated to mimic some environmental disturbance. The training and test dataset sizes are 7,500 and 2,500, respectively. We also evaluated our model under a semi-supervised setting where only 10\% of the annotation vectors are available. Since the annotations are bounded from 0 to 1, CDM's upper and lower bounds are also bounded from 0 to 1. 

\subsubsection{Causally Disentangled Generation}

\begin{figure}[ht]
    \centering
    \subfigure[CausalVAE]{
    \includegraphics[width=\columnwidth]{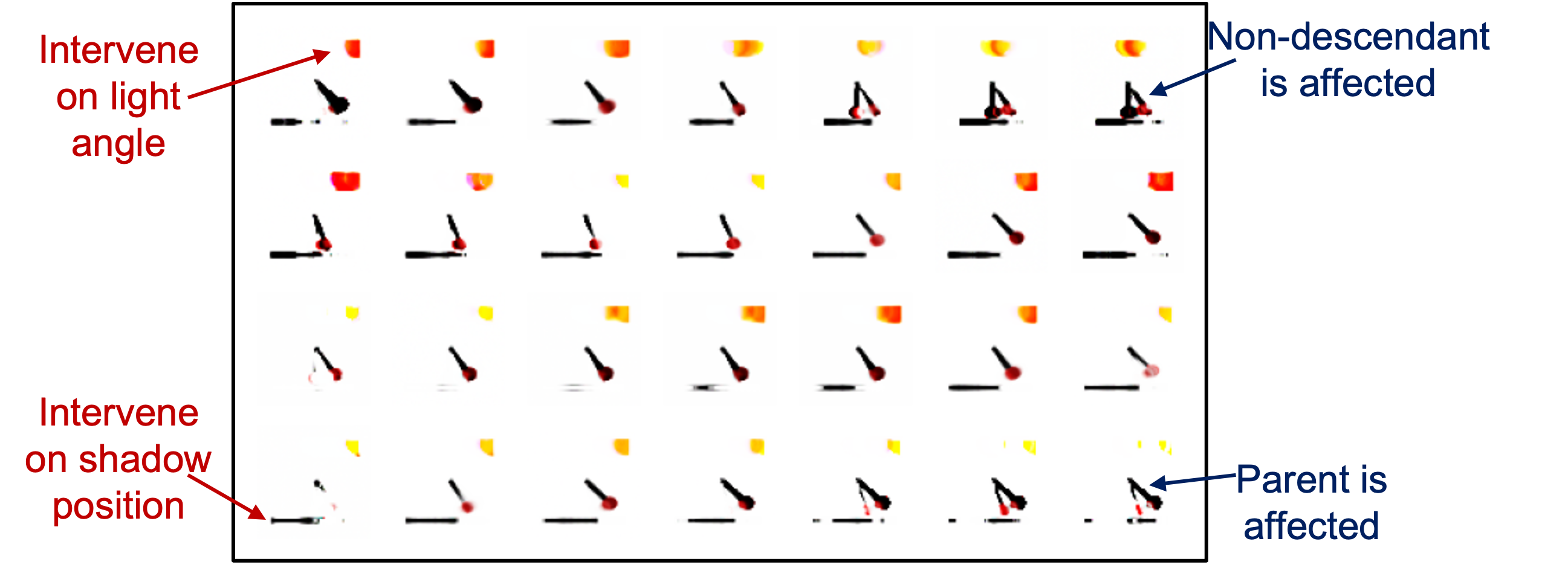} 
    \label{fig:do_causal}}
    
    \subfigure[DEAR]{
    \includegraphics[width=\columnwidth]{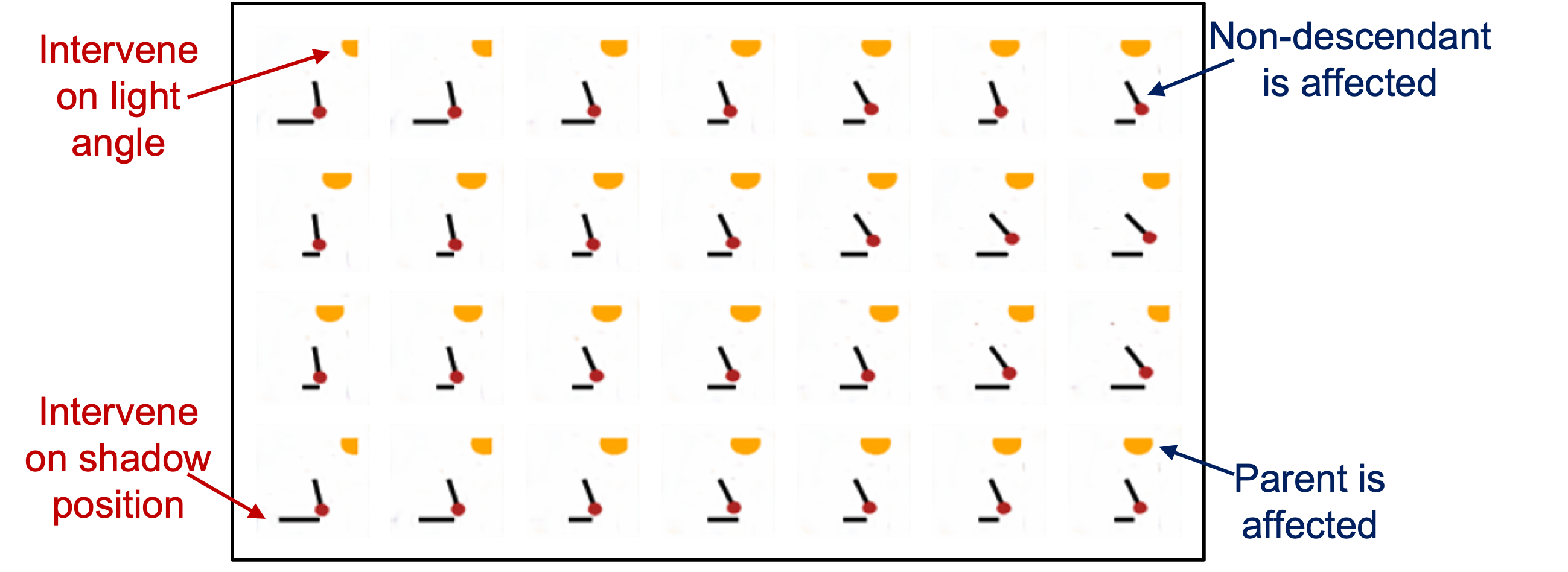} 
    \label{fig:do_dear}}
    
    \subfigure[CDG-VAE(NL)]{
    \includegraphics[width=\columnwidth]{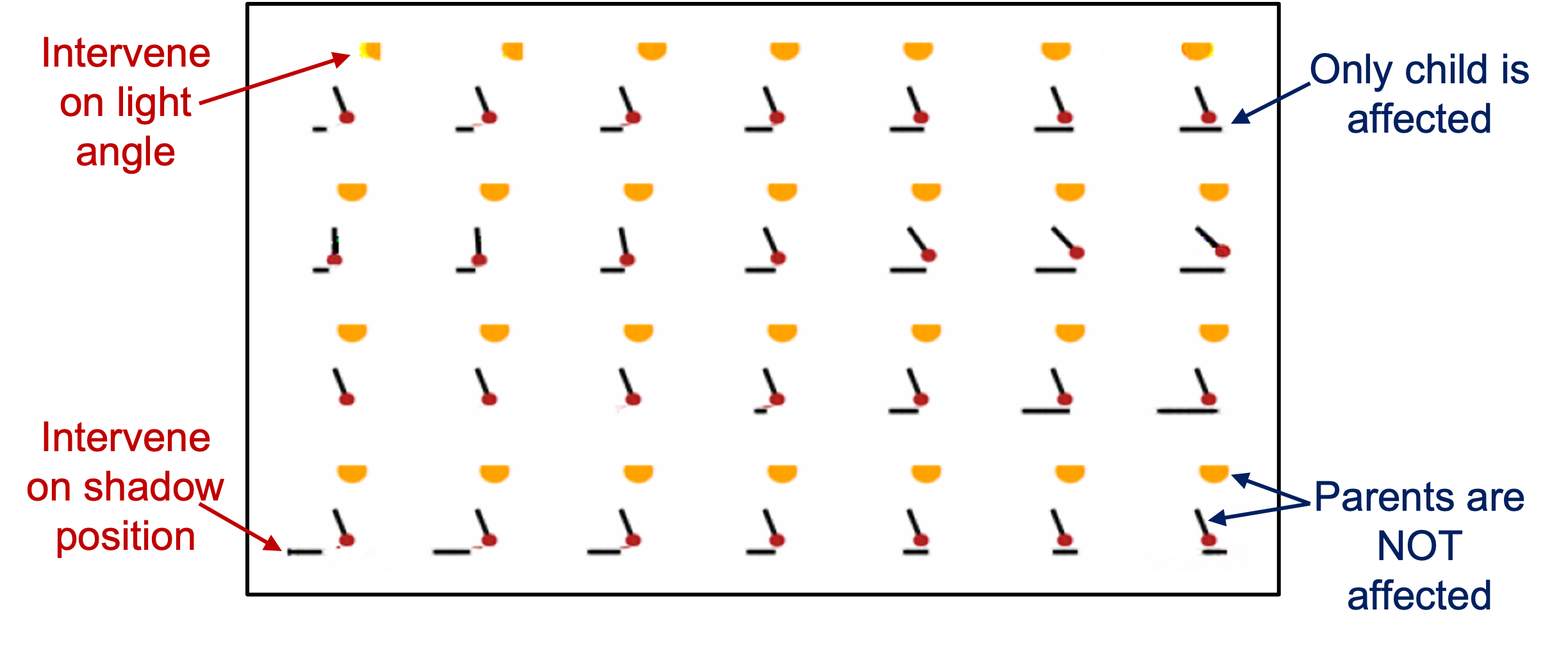} 
    \label{fig:do_our_nonlinear}}

    % \subfigure[CDG-VAE(nonlinear)]{
    % \includegraphics[width=\columnwidth]{fig/do_GAM_nonlinear_overleaf1.png} 
    % \label{fig:do_our_nonlinear}}
    \caption{Examples of generated counterfactual images. For each image, intervened dimensions are light angle, pendulum angle, shadow length, and shadow position from top to bottom. `NL' denotes that a nonlinear $f$ is used.}
    \label{fig:do_result}
\end{figure}

We investigate the performance of counterfactual generation through $do$-intervention on the latent variable, and Figure \ref{fig:do_result} shows corresponding generated images. For CausalVAE and DEAR, if we intervene on shadow length and position, block partitions of their parents (i.e., light angle and pendulum angle) are affected (see the third and fourth row of Figure \ref{fig:do_causal} and \ref{fig:do_dear}). However, CDG-VAE can generate images in which block partitions of light angle and pendulum angle are not affected when their children (i.e., shadow length and position) are intervened (see the third and fourth row of Figure \ref{fig:do_our_nonlinear}). Therefore, CDG-VAE under Proposition \ref{prop:decoder} enables CDG.

% For VAE and InfoMax VAE, even if we $do$-intervene on latent variables of shadow length and position (the third and fourth row of Figure \ref{fig:do_vae}, \ref{fig:do_infomax}), the image factor of shadow length and position in generated images are not changed. Also, f

\subsubsection{Causal Disentanglement Metric}

\begin{table*}[ht]
\caption{Numbers in parentheses are lower and upper bounds of CDM. `L' and `NL' denote the model with linear and nonlinear $f$, and `*' denotes the semi-supervised learned model. Mean and standard deviation values are obtained from 10 repeated experiments. `pos' denotes shadow position. $\uparrow$ denotes higher is better and $\downarrow$ denotes lower is better.}
  \centering
  \begin{tabular}{lrrrr}
    \toprule
 & \multicolumn{2}{c}{Interventional Robustness $\downarrow$} & \multicolumn{2}{c}{Counterfactual Generativeness $\uparrow$} \\
    \midrule
Model & $CDM(light, length)$ & $CDM(angle, pos)$ & $CDM(length, angle)$ & $CDM(pos, pos)$ \\
    \midrule
VAE(L) & $(0.44, 0.44)_{\pm (0.35, 0.35)}$ & $(0.28, 0.28)_{\pm (0.30, 0.31)}$ & $(0.31, 0.32)_{\pm (0.16, 0.15)}$ & $(0.27, 0.28)_{\pm (0.25, 0.24)}$ \\
VAE(NL) & $(0.38, 0.40)_{\pm (0.28, 0.27)}$ & $(0.27, 0.33)_{\pm (0.25, 0.24)}$ & $(0.33, 0.34)_{\pm (0.12, 0.12)}$ & $(0.31, 0.34)_{\pm (0.21, 0.20)}$ \\
InfoMax(L) & $(0.42, 0.43)_{\pm (0.39, 0.38)}$ & $(0.38, 0.38)_{\pm (0.34, 0.34)}$ & $(0.40, 0.40)_{\pm (0.26, 0.25)}$ & $(0.29, 0.31)_{\pm (0.22, 0.20)}$ \\
InfoMax(NL) & $(0.37, 0.39)_{\pm (0.32, 0.30)}$ & $(0.26, 0.33)_{\pm (0.28, 0.25)}$ & $\textbf{(0.44, 0.44)}_{\pm (0.21, 0.21)}$ & $(0.31, 0.34)_{\pm (0.19, 0.16)}$ \\
CausalVAE & $(0.28, 0.28)_{\pm (0.11, 0.10)}$ & $(0.17, 0.17)_{\pm (0.09, 0.08)}$ & $(0.10, 0.10)_{\pm (0.04, 0.04)}$ & $(0.29, 0.29)_{\pm (0.09, 0.09)}$ \\
DEAR & $(0.21, 0.23)_{\pm (0.16, 0.15)}$ & $(0.26, 0.29)_{\pm (0.25, 0.24)}$ & $(0.23, 0.25)_{\pm (0.23, 0.23)}$ & $(0.16, 0.20)_{\pm (0.18, 0.16)}$ \\
CDG-VAE(L) & $\textbf{(0.00, 0.00)}_{\pm (0.00, 0.00)}$ & $\textbf{(0.00, 0.00)}_{\pm (0.00, 0.00)}$ & $(0.24, 0.25)_{\pm (0.10, 0.09)}$ & $(0.69, 0.69)_{\pm (0.25, 0.25)}$ \\
CDG-VAE(NL) & $\textbf{(0.00, 0.00)}_{\pm (0.00, 0.00)}$ & $\textbf{(0.00, 0.00)}_{\pm (0.00, 0.00)}$ & $(0.35, 0.36)_{\pm (0.16, 0.15)}$ & $\textbf{(0.78, 0.78)}_{\pm (0.24, 0.24)}$ \\
    \midrule
CDG-VAE(L)* & $\textbf{(0.00, 0.00)}_{\pm (0.00, 0.00)}$ & $\textbf{(0.00, 0.00)}_{\pm (0.00, 0.00)}$ & $(0.21, 0.22)_{\pm (0.09, 0.07)}$ & $(0.66, 0.66)_{\pm (0.22, 0.22)}$ \\
CDG-VAE(NL)* & $\textbf{(0.00, 0.00)}_{\pm (0.00, 0.00)}$ & $\textbf{(0.00, 0.00)}_{\pm (0.00, 0.00)}$ & $(0.29, 0.30)_{\pm (0.12, 0.11)}$ & $\textbf{(0.79, 0.79)}_{\pm (0.21, 0.21)}$ \\
    \bottomrule
  \end{tabular}
\label{tab:cdm}
\end{table*}

We compared models using our proposed causal disentanglement metric CDM. For ease of explanation, we use $CDM(light, length)$ instead of $CDM(c=1, i=3)$. The second and third columns of Table \ref{tab:cdm} indicate the interventional robustness. For example, as light angle is a parent of shadow length, $CDM(light, length)$ measures interventional robustness. CDG-VAE achieves exactly zero in both $CDM(light, length)$ and $CDM(angle, pos)$, which implies that only CDG-VAE can satisfy CDG. It is worth mentioning that CDG-VAE achieves exactly zero CDM values for all other cases (see Appendix A.8). 
% However, the CDM values of other models for both cases are greater than zero, which indicates that other models do not satisfy Definition \ref{def:CDG}. 

And the fourth and fifth columns of Table \ref{tab:cdm} show the counterfactual generativeness. For example, since shadow length is a child of pendulum angle, $CDM(length, angle)$ measures the counterfactual generativeness. CDG-VAE achieves the highest score in $CDM(pos, pos)$ and outperforms CausalVAE and DEAR models in $CDM(length, angle)$. Therefore, Table \ref{tab:cdm} indicates that CDG-VAE has competitive counterfactual generativeness performance (see Appendix A.8 for other cases). 

\subsubsection{Downstream Task}

% The previous experiment results verify that our CDG-VAE has a competitive visual and causal disentanglement generation metric performance. 
This section investigates the advantages of causally disentangled representations for two downstream tasks: sample efficiency and distributional robustness \cite{shen2022weakly}. The binary classification task is mainly used for evaluation, and we generate the target label as a function of the ground-truth factors. It means that the representations learned from the generative model are causal representations of the target label (see Appendix A.7 for a detailed explanation).

\textbf{Sample Efficiency.} To measure the sample efficiency, we use the statistical efficiency score defined as the test accuracy based on 100 samples divided by the test accuracy based on all samples, following \cite{locatello2019challenging, shen2022weakly}. We use fitted encoders to extract representations and train an MLP classifier on top of the representations to predict the target label. Notably, all models are evaluated with the same MLP. We also report the test accuracies to prevent misleading when a classifier achieves poor test accuracy in both cases. Table \ref{tab:se} shows that CDG-VAE performs the best in the sample efficiency downstream task. 

\begin{table}
\caption{Sample efficiency and test accuracies with 100 and all training samples. `L' and `NL' denote the model with linear and nonlinear $f$, and `*' denotes the semi-supervised learned model. Mean and standard deviation values are obtained from 10 repeated experiments. Higher is better.}
  \centering
    \resizebox{\columnwidth}{!}{
  \begin{tabular}{lrrr}
    \toprule
    Model & 100(\%) & All(\%) & SE(100/All) \\
    \midrule
VAE(L) & $88.41_{\pm 2.07}$ & $90.18_{\pm 1.01}$ & $98.03_{\pm 1.67}$ \\
VAE(NL) & $87.74_{\pm 1.23}$ & $90.16_{\pm 0.60}$ & $97.32_{\pm 1.25}$ \\
InfoMax(L) & $88.69_{\pm 1.36}$ & $90.34_{\pm 0.52}$ & $98.18_{\pm 1.45}$ \\
InfoMax(NL) & $87.74_{\pm 1.05}$ & $90.15_{\pm 0.54}$ & $97.32_{\pm 0.98}$ \\
CausalVAE & $50.39_{\pm 2.08}$ & $86.94_{\pm 1.44}$ & $57.98_{\pm 2.74}$ \\
DEAR & $82.92_{\pm 3.45}$ & $88.92_{\pm 1.34}$ & $93.27_{\pm 3.97}$ \\
CDG-VAE(L) & $89.48_{\pm 0.76}$ & $90.68_{\pm 0.24}$ & $\textbf{98.67}_{\pm 0.81}$ \\
CDG-VAE(NL) & $87.94_{\pm 1.29}$ & $90.19_{\pm 0.52}$ & $97.51_{\pm 1.22}$ \\
\midrule
CDG-VAE(L)* & $89.33_{\pm 0.63}$ & $90.52_{\pm 0.34}$ & $\textbf{98.69}_{\pm 0.48}$ \\
CDG-VAE(NL)* & $88.39_{\pm 0.69}$ & $90.06_{\pm 0.47}$ & $98.14_{\pm 0.90}$ \\
    \bottomrule
  \end{tabular}}
\label{tab:se}
\end{table}

\textbf{Distributional Robustness.} To evaluate the distributional robustness of causal representations, we manipulate the training dataset to impose spurious correlations between the target label and some spurious features of the image. We choose $background\_color \in \{white (0), blue (1)\}$ as a spurious feature \cite{shen2022weakly}. 80\% of the training samples have the same value of the target label and $background\_color$ (strong correlation), but the test samples do not have such correlations (all label values are distributed equally).

Table \ref{tab:dr} shows the performances of the compared models in the distributional robustness (downstream task) of the causally disentangled representation. The worst case (`TrainWorst' and `TestWorst') is when the target label and the spurious feature $background\_color$ are grouped to have the opposite label. For that group, the spurious feature has a different correlation between the training and test datasets. And Table \ref{tab:dr} shows that CDG-VAE shows the best test accuracy in the worst cases. Therefore, CDG-VAE can produce a causally disentangled representation robust to distributional shifts. Note that `TrainAvg' or `TrainWorst' are not reasonable criteria to judge the best model because Table \ref{tab:dr} shows the distributional robustness where the distribution of the test dataset is changed (i.e., distributional shift).

On the other hand, if the latent space is not causally disentangled, the features of observation partitions are entangled in the encoder, and the model exploits the entangled features in the learning of the latent space. It results that the latent variable is affected by the changes in the observation partitions, which are not causally related. See the toy example of such a case in Appendix A.7. And we guess that the latent space of CausalVAE is not fully disentangled, and the downstream classifier utilizes the correlation information between the spurious feature and the target label. Consequently, the downstream classifier with CausalVAE is overfitted and shows a higher score in `TrainAvg' metric in Table \ref{tab:dr}.

\begin{table*}[ht]
\caption{Distributional robustness: Train and test dataset accuracy for average (`Avg') and worst (`Worst') cases. `L' and `NL' denote the model with linear and nonlinear $f$, and `*' denotes the semi-supervised learned model. Mean and standard deviation values are obtained from 10 repeated experiments. Higher is better.}
\label{tab:dr}
  \centering
  \begin{tabular}{lrrrr}
    \toprule
    Model & TrainAvg(\%) & TrainWorst(\%) & TestAvg(\%) & TestWorst(\%) \\
    \midrule
VAE(L) & $61.70_{\pm 2.46}$ & $57.70_{\pm 2.56}$ & $58.88_{\pm 0.85}$ & $55.73_{\pm 2.75}$ \\
VAE(NL) & $62.14_{\pm 1.98}$ & $57.63_{\pm 2.93}$ & $59.41_{\pm 0.93}$ & $55.91_{\pm 3.34}$ \\
InfoMax(L) & $61.50_{\pm 2.08}$ & $57.55_{\pm 3.56}$ & $59.04_{\pm 1.21}$ & $56.00_{\pm 3.29}$ \\
InfoMax(NL) & $62.33_{\pm 2.35}$ & $58.07_{\pm 2.48}$ & $\textbf{59.60}_{\pm 0.77}$ & $56.45_{\pm 2.36}$ \\
CausalVAE & $\textbf{73.93}_{\pm 0.99}$ & $35.52_{\pm 5.57}$ & $57.92_{\pm 1.28}$ & $33.91_{\pm 4.92}$ \\
DEAR & $62.33_{\pm 2.27}$ & $55.62_{\pm 4.61}$ & $58.60_{\pm 1.02}$ & $53.16_{\pm 4.25}$ \\
CDG-VAE(L) & $64.28_{\pm 5.22}$ & $50.45_{\pm 10.07}$ & $58.02_{\pm 0.91}$ & $48.59_{\pm 10.05}$ \\
CDG-VAE(NL) & $60.97_{\pm 0.91}$ & $\textbf{59.34}_{\pm 1.30}$ & $59.22_{\pm 0.60}$ & $\textbf{57.21}_{\pm 1.50}$ \\
\midrule
CDG-VAE(L)* & $70.43_{\pm 4.69}$ & $40.47_{\pm 11.16}$ & $55.36_{\pm 1.37}$ & $36.31_{\pm 9.87}$ \\
CDG-VAE(NL)* & $67.19_{\pm 3.61}$ & $44.28_{\pm 9.88}$ & $55.66_{\pm 1.96}$ & $41.01_{\pm 9.48}$ \\
    \bottomrule
  \end{tabular}
\end{table*}

\subsection{Tabular Datasets}
\label{sec:4.3}

\begin{table*}[ht]
\caption{SHD and data quality scores for synthetic datasets. All models use linear $f$. Mean and standard deviation values are obtained from 10 repeated experiments. `Baseline' indicates results from the observed dataset. $\uparrow$ denotes higher is better and $\downarrow$ denotes lower is better.}
\label{tab:tabular}
  \centering
  \begin{tabular}{lrrrrrr}
    \toprule
    Dataset & \multicolumn{2}{c}{\texttt{loan}$(p=5)$} & \multicolumn{2}{c}{\texttt{adult}$(p=5)$} & \multicolumn{2}{c}{\texttt{covertype}$(p=8)$} \\
    % Dataset & \multicolumn{2}{c}{\texttt{loan} ($p=5$)} & \multicolumn{2}{c}{\texttt{adult} ($p=5$)} & \multicolumn{2}{c}{\texttt{covertype} ($p=8$)} \\
    \midrule
    Model & SHD $\downarrow$ & $R^2$ $\uparrow$ & SHD $\downarrow$ & $F_1$ $\uparrow$ & SHD $\downarrow$ & $F_1$ $\uparrow$ \\
    \midrule
Baseline & - & $0.392$ & - & $0.818$ & - & $0.712$\\
VAE & $6.1_{\pm 2.3}$ & $-7.936_{\pm 15.127}$ & $7.0_{\pm 2.3}$ & $0.739_{\pm 0.032}$ & $17.7_{\pm 3.1}$ & $0.067_{\pm 0.022}$\\
InfoMax & $7.1_{\pm 1.1}$ & $-5.149_{\pm 7.920}$ & $7.5_{\pm 1.2}$ & $0.712_{\pm 0.061}$ & $17.8_{\pm 4.1}$ & $0.102_{\pm 0.021}$\\
TVAE & $5.0_{\pm 1.8}$ & $-0.631_{\pm 0.380}$ & $5.1_{\pm 1.5}$ & $0.724_{\pm 0.009}$ & $16.2_{\pm 2.9}$ & $\textbf{0.358}_{\pm 0.024}$\\
CTAB-GAN & $4.9_{\pm 0.9}$ & $-0.912_{\pm 0.504}$ & $5.5_{\pm 3.2}$ & $\textbf{0.795}_{\pm 0.007}$ & $17.3_{\pm 3.6}$ & $0.077_{\pm 0.029}$\\
    \midrule
CDG-VAE & $0.9_{\pm 0.3}$ & $-0.982_{\pm 1.663}$ & $0.3_{\pm 0.9}$ & $0.696_{\pm 0.003}$ & $\textbf{1.6}_{\pm 0.5}$ & $0.127_{\pm 0.015}$\\
CDG-TVAE & $\textbf{0.4}_{\pm 0.5}$ & $\textbf{0.013}_{\pm 0.010}$ & $\textbf{0.2}_{\pm 0.4}$ & $0.645_{\pm 0.001}$ & $2.8_{\pm 0.6}$ & $0.178_{\pm 0.005}$\\
    \bottomrule
  \end{tabular}
\end{table*}

\subsubsection{Causal Disentanglement Learning}

\begin{figure}
    \centering
    \subfigure[]{\resizebox{0.35\columnwidth}{!}{
    \begin{tikzpicture}[shorten >=1pt,auto,thick,main node/.style={circle,fill=blue!20,draw,font=\sffamily\Large\bfseries}]
        % nodes 
        \node[state] (g1) {$\bg_1$};
        \node[state] (g2) [right = 1.4cm of g1] {$\bg_2$};
        \node[state] (g3) [below right = 0.7cm and 0.6cm of g1] {$\bg_3$};
    
        % Directed edge
        \path (g1) edge (g3);
        \path (g2) edge (g3);
    \end{tikzpicture} \label{fig:dag_loan1}}}
    \subfigure[]{\includegraphics[width=0.53\columnwidth]{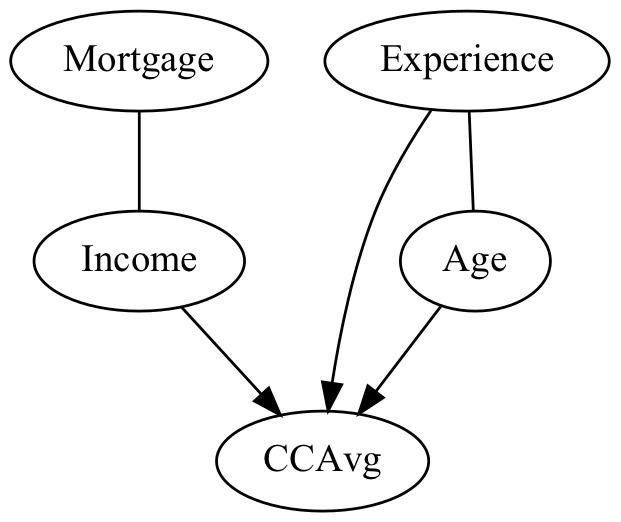} \label{fig:dag_loan2}}
    % \subfigure[]{\includegraphics[width=0.3\columnwidth]{fig/dag_bijection_loan.png} \label{fig:dag_loan2}}
    \caption{\texttt{loan} dataset. (a) DAG structure of ground-truth factors $\bg$. (b) The chain graph of covariates.}
    \label{fig:dag_loan}
\end{figure}
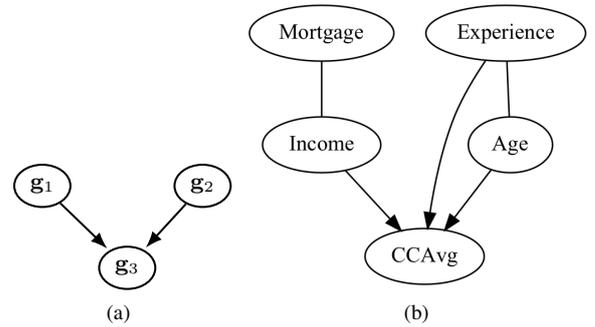

In the supervised causal disentanglement learning method with tabular datasets, we assume that the causal dependencies between $K$ block subvectors of $\bx_{\sigma}$ are induced by the causal structure of $\bg_{\pi}$. Therefore, $\bx_{\sigma}$ is a chain graph-structured data, such as a multi-layered proteomic data \cite{ha2021bayesian} and $\bx_{\sigma(1)},\cdots,\bx_{\sigma(K)}$ are chain components \cite{koller2009probabilistic}.

Figure \ref{fig:dag_loan2} shows the chain graph of \texttt{loan} dataset obtained by PC algorithm \cite{spirtes2000causation}. The chain components of $\bx_\sigma$ are $\bx_{\sigma(1)} = (\mbox{Mortgage, Income}), \bx_{\sigma(2)} = (\mbox{Experiences, Age})$, and $\bx_{\sigma(3)} = (\mbox{CCAvg})$. We assume that each chain component $\bx_{\sigma(j)}$ is generated by the single ground-truth factor $\bg_j$, as in GM3 of Figure \ref{fig:generative} ($\bg_j \rightarrow \bx_{\sigma(j)}$). Thus, Figure \ref{fig:dag_loan1} is the DAG structure of the ground-truth factors $\bg$.
% Note that relationships including undirected edges between variables of a single component $\bx_{\sigma(j)}, j=1,\cdots, K$ are not modeled. 
% and the causal topological order of $\bg$ is $(1, 2, 3)$ (or $(2, 1, 3)$) 
% and let $\bx = (\bx_1, \cdots, \bx_5)$ denote the covariates Mortgage, Income, Experience, Age, and CCAvg, in order. 
% So, chain components are $\bx_{\sigma(1)} = \{\bx_1, \bx_2\}, \bx_{\sigma(2)} = \{\bx_3, \bx_4\}$, and $\bx_{\sigma(3)} = \{\bx_5\}$. 

Due to undirected edges between covariates (e.g., the edge between Mortgage and Income), the SCM of covariates is not defined well, and the covariate-wise topological generation of \cite{xu2019achieving, van2021decaf, wen2021causal} is not applicable. However, without a completely identified DAG, CDG-VAE can achieve the CDG. That is, our model can include (Mortgage, Income) and (Experience, Age) as the chain components in disentanglement learning.
% requires a well-defined SCM of latent variables and labels to disentangle the latent space.

First, since Figure \ref{fig:dag_loan1} consists of only directed causal relationships, the SCM of the latent variables can be formulated based on the DAG of Figure \ref{fig:dag_loan1}. Next, we define the bijection output of each chain component $\bx_{\sigma(j)}$ as $\mbE[\bu_j|\bx_{\sigma(j)}]$ and assume that $\bg_j = \mbE[\bu_j|\bx_{\sigma(j)}] = \mbE[\bu_j|\bx]$, for $j=1,2,3$. The DAG obtained by the PC algorithm with bijection outputs is equivalent to Figure \ref{fig:dag_loan1}, implying that bijection outputs have the same causal structure as the ground-truth factors. In practice, we use the interleaving function for bijection after the min-max scaling (note that the independence is not affected by scaling). See Appendix A.6 for other tabular datasets' chain graph structures and chain components.
% If the size of a block partition indices set is 1, then the bijection is identity.

% and bijection are used as the labels of supervised regularization. 
% Then, the joint distribution is factorized as
% \bea \label{eq:loan2}
% p(\bu_1, \bu_2, \bu_3) = p(\bu_3| \bu_1, \bu_2) p(\bu_1) p(\bu_2).
% \eea

% Regardless of the ground-truth direction of undirected edges between $(\bx_1, \bx_2)$ and $(\bx_3, \bx_4)$, $\bx_1, \cdots, \bx_4$ must precede $\bx_5$ in the causal topological order. So, 
% So, the joint distribution is factorized as 
% \bea \label{eq:loan1}
% p(\bx_1, \cdots, \bx_5) = p(\bx_{\sigma(3)}| \bx_{\sigma(1)}, \bx_{\sigma(2)}) p(\bx_{\sigma(1)}) p(\bx_{\sigma(2)}).
% \eea

\subsubsection{Synthetic Data Generation}

We evaluated the performance of our model in synthetic data generation.
% First, we assume there is no unmeasured confounder. 
To measure whether the observed causal structure is preserved, we use the Structural Hamming Distance (SHD) \cite{tsamardinos2006max} between causal graphs of the observed and a synthetic dataset. Causal graphs are obtained by the PC algorithm \cite{spirtes2000causation}. The smaller SHD value indicates a model can generate synthetic data with precise causal relationship information.
% The smaller SHD value indicates that the causal relationship information of a synthetic dataset is precise. 

On the other hand, to evaluate the synthetic data quality, we use the synthetic data as training data for three widely used machine learning algorithms: linear (logistic) regression, Random Forest, and Gradient Boosting (see Appendix A.6 for details). We average the following metrics: $R^2$ for the regression and $F_1$ for the classification problem. Note that the synthetic data and the observed data have the same size. Here, CDG-TVAE is the new synthesizer where the mode-specific normalization technique of TVAE is combined with CDG-VAE.

The SHD values of Table \ref{tab:tabular} show that models under Proposition \ref{prop:decoder} (CDG-VAE and CDG-TVAE) outperform in terms of preserving the original causal structure well compared to the other models in synthetic data generation. Moreover, we observe that the difference in the SHD value between the proposed model and the compared models becomes significantly larger as the number of covariates ($p$) increases. 

Although CDG-TVAE compromises the $F_1$ score and SHD value (\texttt{covertype} dataset) since we restrict the latent space and the decoder structure of the model according to the causal relationships, CDG-TVAE has a competitive performance in data quality scores in Table \ref{tab:tabular} (\texttt{loan}, \texttt{adult} datasets). Therefore, CDG-TVAE can generate synthetic data, which can be used as a good proxy of the original data while preserving the observed causal structure.

\section{Conclusion and Limitations}
\label{sec:5}

% This paper proposes a new supervised learning method for VAE to achieve causally disentangled generation (CDG), which is newly defined in this paper. The CDG means generating data under causal mechanisms described by a given structural causal model (SCM), and we show that the CDG depends on both the encoder and decoder of VAE. To enable CDG, we adapt supervised regularization of the encoder and prove the sufficient condition of the decoder. Furthermore, we devise metrics for evaluating causal disentanglement based on the necessary conditions for CDG.

We demonstrate that causally disentangled latent space is insufficient to generate an image according to the causal mechanism defined by a general SCM. Figure \ref{fig:do_result} shows that the entangled decoder can not generate the counterfactual image even when the causal structure is simple and the block-partition indices are known. This result concludes that the disentanglement of features is crucial both in the encoder and the decoder, and a causal relationship is contaminated without the disentanglement.

The decoder structure of Proposition \ref{prop:decoder} is very restricted since it requires the indices of all partition blocks (i.e., obtaining complete causal information for supervision). We do not believe the proposed decoder structure is directly applicable to large-scale datasets such as the CelebA dataset \cite{liu2015faceattributes}. Besides, in synthetic data generation, we numerically find that the supervised disentanglement learning with interleaving bijection (Section \ref{sec:4.3}) can preserve the observed causal structure only when the cardinality of a chain component is small (empirically, less than 4). We guess that it is because the interleaving function has limited expressiveness. 
% Despite the limitation, our results show that the disentanglement in the decoder is critical in causal disentanglement learning and provides a new research topic, such as fine-tuning a decoder with the group LASSO penalty. 
% Also, we think that it may be possible to learn the automatic block-partitioning of the image with segmentation techniques, but we can not reach the ideal generative model not yet.

We expect that we can effectively deal with the partitioned blocks as a multi-layered image with the same size and control the property of disentangled features by utilizing mutual information between generated samples and annotation vectors. In addition, to enrich the expressiveness of the latent space with a causal structure, it is necessary to find computationally tractable bijections maps for multivariates and multi-layered data and apply supervised disentanglement learning. We leave solving the two problems as our future work.

Lastly, our model is identifiable in the sense of \cite{Kivva2022IdentifiabilityOD} (the generalized identifiability). But it is not sure that our model is also $A$-identifiable or $P$-indentifiable \cite{Khemakhem2019VariationalAA} because our standard Gaussian assumption for priors violates the sufficient condition of the identifiability. We will thoroughly discuss the identifiability of our proposed model and address this as a crucial research topic for future investigations.

% Our proposed metric CDM has a limitation in requiring accessible labels to cover sufficient data distribution as \cite{shen2022weakly} referred. And CDM uses only necessary conditions to assess a generative model's compliance with CDG, which can limit the interpretability of the metric's results. 
% In the image domain, partition tuples are assumed to be static. We leave the development of CDG-VAE in time-series causality datasets \cite{yi2019clevrer, ahmed2020causalworld, mcduff2022causalcity} as our future work. 
% We leave the development of CDG-VAE in the fair dataset generation.

\ack 
This work was supported by the National Research Foundation of Korea(NRF) grant funded by the Korea government(MSIT) (No. NRF-2022R1A4A3033874, No. NRF-2022R1F1A1074758, and NO.2022M3J6A1084845). The authors acknowledge the Urban Big data and AI Institute of the University of Seoul supercomputing resources (\url{http://ubai.uos.ac.kr}) made available for conducting the research reported in this paper.
% J. Jeon was partly supported by the National Research Foundation of Korea grant 2022R1A4A3033874 and 2022R1F1A1074758.
% This research was also supported by the National Research Foundation of Korea(NRF) grant funded by the Korea government(MSIT) (NO.2022M3J6A1084845). 

\bibliography{ecai}

\begin{thebibliography}{10}

\bibitem{An2023CustomizedLS}
SeungHwan An and Jong-June Jeon, `Customized latent space: Practical usage of
  variational autoencoder', {\em SSRN Electronic Journal}, (2023).

\bibitem{behrmann2019invertible}
Jens Behrmann, Will Grathwohl, Ricky~TQ Chen, David Duvenaud, and
  J{\"o}rn-Henrik Jacobsen, `Invertible residual networks', in {\em
  International Conference on Machine Learning}, pp. 573--582. PMLR, (2019).

\bibitem{bengio2013representation}
Yoshua Bengio, Aaron Courville, and Pascal Vincent, `Representation learning: A
  review and new perspectives', {\em IEEE transactions on pattern analysis and
  machine intelligence}, {\bf 35}(8),  1798--1828, (2013).

\bibitem{bengio2019meta}
Yoshua Bengio, Tristan Deleu, Nasim Rahaman, Rosemary Ke, S{\'e}bastien
  Lachapelle, Olexa Bilaniuk, Anirudh Goyal, and Christopher Pal, `A
  meta-transfer objective for learning to disentangle causal mechanisms', {\em
  arXiv preprint arXiv:1901.10912}, (2019).

\bibitem{bouchacourt2018multi}
Diane Bouchacourt, Ryota Tomioka, and Sebastian Nowozin, `Multi-level
  variational autoencoder: Learning disentangled representations from grouped
  observations', in {\em Proceedings of the AAAI Conference on Artificial
  Intelligence}, volume~32, (2018).

\bibitem{Burda2015ImportanceWA}
Yuri Burda, Roger~Baker Grosse, and Ruslan Salakhutdinov, `Importance weighted
  autoencoders', {\em CoRR}, {\bf abs/1509.00519}, (2015).

\bibitem{dupont2018learning}
Emilien Dupont, `Learning disentangled joint continuous and discrete
  representations', {\em Advances in Neural Information Processing Systems},
  {\bf 31}, (2018).

\bibitem{goodfellow2020generative}
Ian Goodfellow, Jean Pouget-Abadie, Mehdi Mirza, Bing Xu, David Warde-Farley,
  Sherjil Ozair, Aaron Courville, and Yoshua Bengio, `Generative adversarial
  networks', {\em Communications of the ACM}, {\bf 63}(11),  139--144, (2020).

\bibitem{ha2021bayesian}
Min~Jin Ha, Francesco~Claudio Stingo, and Veerabhadran Baladandayuthapani,
  `Bayesian structure learning in multilayered genomic networks', {\em Journal
  of the American Statistical Association}, {\bf 116}(534),  605--618, (2021).

\bibitem{Hajimiri2021SemiSupervisedDO}
Sina Hajimiri, Aryo Lotfi, and Mahdieh~Soleymani Baghshah, `Semi-supervised
  disentanglement of class-related and class-independent factors in vae', {\em
  ArXiv}, {\bf abs/2102.00892}, (2021).

\bibitem{higgins2016beta}
Irina Higgins, Loic Matthey, Arka Pal, Christopher Burgess, Xavier Glorot,
  Matthew Botvinick, Shakir Mohamed, and Alexander Lerchner, `beta-vae:
  Learning basic visual concepts with a constrained variational framework',
  (2016).

\bibitem{Khemakhem2019VariationalAA}
Ilyes Khemakhem, Diederik~P. Kingma, and Aapo Hyv{\"a}rinen, `Variational
  autoencoders and nonlinear ica: A unifying framework', in {\em International
  Conference on Artificial Intelligence and Statistics}, (2019).

\bibitem{Kingma2014SemisupervisedLW}
Diederik~P. Kingma, Shakir Mohamed, Danilo~Jimenez Rezende, and Max Welling,
  `Semi-supervised learning with deep generative models', in {\em NIPS},
  (2014).

\bibitem{kingma2013auto}
Diederik~P Kingma and Max Welling, `Auto-encoding variational bayes', {\em
  arXiv preprint arXiv:1312.6114}, (2013).

\bibitem{Kivva2022IdentifiabilityOD}
Bohdan Kivva, Goutham Rajendran, Pradeep Ravikumar, and Bryon Aragam,
  `Identifiability of deep generative models without auxiliary information', in
  {\em Neural Information Processing Systems}, (2022).

\bibitem{koller2009probabilistic}
Daphne Koller and Nir Friedman, {\em Probabilistic graphical models: principles
  and techniques}, MIT press, 2009.

\bibitem{lachapelle2022disentanglement}
S{\'e}bastien Lachapelle, Pau Rodriguez, Yash Sharma, Katie~E Everett, R{\'e}mi
  Le~Priol, Alexandre Lacoste, and Simon Lacoste-Julien, `Disentanglement via
  mechanism sparsity regularization: A new principle for nonlinear ica', in
  {\em Conference on Causal Learning and Reasoning}, pp. 428--484. PMLR,
  (2022).

\bibitem{liu2015faceattributes}
Ziwei Liu, Ping Luo, Xiaogang Wang, and Xiaoou Tang, `Deep learning face
  attributes in the wild', in {\em Proceedings of International Conference on
  Computer Vision (ICCV)}, (December 2015).

\bibitem{locatello2019challenging}
Francesco Locatello, Stefan Bauer, Mario Lucic, Gunnar Raetsch, Sylvain Gelly,
  Bernhard Sch{\"o}lkopf, and Olivier Bachem, `Challenging common assumptions
  in the unsupervised learning of disentangled representations', in {\em
  international conference on machine learning}, pp. 4114--4124. PMLR, (2019).

\bibitem{locatello2019disentangling}
Francesco Locatello, Michael Tschannen, Stefan Bauer, Gunnar R{\"a}tsch,
  Bernhard Sch{\"o}lkopf, and Olivier Bachem, `Disentangling factors of
  variation using few labels', {\em arXiv preprint arXiv:1905.01258}, (2019).

\bibitem{Locatello2020Disentangling}
Francesco Locatello, Michael Tschannen, Stefan Bauer, Gunnar Rätsch, Bernhard
  Schölkopf, and Olivier Bachem, `Disentangling factors of variations using
  few labels', in {\em International Conference on Learning Representations},
  (2020).

\bibitem{pearl2009causality}
Judea Pearl, {\em Causality}, Cambridge university press, 2009.

\bibitem{Peters2017ElementsOC}
J.~Peters, Dominik Janzing, and Bernhard Sch{\"o}lkopf, `Elements of causal
  inference: Foundations and learning algorithms', (2017).

\bibitem{peters2017elements}
Jonas Peters, Dominik Janzing, and Bernhard Sch{\"o}lkopf, {\em Elements of
  causal inference: foundations and learning algorithms}, The MIT Press, 2017.

\bibitem{reddy2022causally}
Abbavaram~Gowtham Reddy, Vineeth~N Balasubramanian, et~al., `On causally
  disentangled representations', in {\em Proceedings of the AAAI Conference on
  Artificial Intelligence}, volume~36, pp. 8089--8097, (2022).

\bibitem{rezaabad2020learning}
Ali~Lotfi Rezaabad and Sriram Vishwanath, `Learning representations by
  maximizing mutual information in variational autoencoders', in {\em 2020 IEEE
  International Symposium on Information Theory (ISIT)}, pp. 2729--2734. IEEE,
  (2020).

\bibitem{rezende2015variational}
Danilo Rezende and Shakir Mohamed, `Variational inference with normalizing
  flows', in {\em International conference on machine learning}, pp.
  1530--1538. PMLR, (2015).

\bibitem{shen2022weakly}
Xinwei Shen, Furui Liu, Hanze Dong, Qing Lian, Zhitang Chen, and Tong Zhang,
  `Weakly supervised disentangled generative causal representation learning',
  {\em Journal of Machine Learning Research}, {\bf 23},  1--55, (2022).

\bibitem{Sicks2020AGL}
Robert Sicks, Ralf Korn, and Stefanie Schwaar, `A generalised linear model
  framework for variational autoencoders based on exponential dispersion
  families', {\em J. Mach. Learn. Res.}, {\bf 22},  233:1--233:41, (2020).

\bibitem{spirtes2000causation}
Peter Spirtes, Clark~N Glymour, Richard Scheines, and David Heckerman, {\em
  Causation, prediction, and search}, MIT press, 2000.

\bibitem{suter2019robustly}
Raphael Suter, Djordje Miladinovic, Bernhard Sch{\"o}lkopf, and Stefan Bauer,
  `Robustly disentangled causal mechanisms: Validating deep representations for
  interventional robustness', in {\em International Conference on Machine
  Learning}, pp. 6056--6065. PMLR, (2019).

\bibitem{szabo2017challenges}
Attila Szab{\'o}, Qiyang Hu, Tiziano Portenier, Matthias Zwicker, and Paolo
  Favaro, `Challenges in disentangling independent factors of variation', {\em
  arXiv preprint arXiv:1711.02245}, (2017).

\bibitem{trauble2021disentangled}
Frederik Tr{\"a}uble, Elliot Creager, Niki Kilbertus, Francesco Locatello,
  Andrea Dittadi, Anirudh Goyal, Bernhard Sch{\"o}lkopf, and Stefan Bauer, `On
  disentangled representations learned from correlated data', in {\em
  International Conference on Machine Learning}, pp. 10401--10412. PMLR,
  (2021).

\bibitem{tsamardinos2006max}
Ioannis Tsamardinos, Laura~E Brown, and Constantin~F Aliferis, `The max-min
  hill-climbing bayesian network structure learning algorithm', {\em Machine
  learning}, {\bf 65}(1),  31--78, (2006).

\bibitem{van2021decaf}
Boris van Breugel, Trent Kyono, Jeroen Berrevoets, and Mihaela van~der Schaar,
  `Decaf: Generating fair synthetic data using causally-aware generative
  networks', {\em Advances in Neural Information Processing Systems}, {\bf 34},
   22221--22233, (2021).

\bibitem{wen2021causal}
Bingyang Wen, Luis~Oliveros Colon, KP~Subbalakshmi, and Rajarathnam
  Chandramouli, `Causal-tgan: Generating tabular data using causal generative
  adversarial networks', {\em arXiv preprint arXiv:2104.10680}, (2021).

\bibitem{xu2019achieving}
Depeng Xu, Yongkai Wu, Shuhan Yuan, Lu~Zhang, and Xintao Wu, `Achieving causal
  fairness through generative adversarial networks', in {\em Proceedings of the
  Twenty-Eighth International Joint Conference on Artificial Intelligence},
  (2019).

\bibitem{xu2019modeling}
Lei Xu, Maria Skoularidou, Alfredo Cuesta-Infante, and Kalyan Veeramachaneni,
  `Modeling tabular data using conditional gan', {\em Advances in Neural
  Information Processing Systems}, {\bf 32}, (2019).

\bibitem{yang2021causalvae}
Mengyue Yang, Furui Liu, Zhitang Chen, Xinwei Shen, Jianye Hao, and Jun Wang,
  `Causalvae: Disentangled representation learning via neural structural causal
  models', in {\em Proceedings of the IEEE/CVF Conference on Computer Vision
  and Pattern Recognition}, pp. 9593--9602, (2021).

\bibitem{yu2019dag}
Yue Yu, Jie Chen, Tian Gao, and Mo~Yu, `Dag-gnn: Dag structure learning with
  graph neural networks', in {\em International Conference on Machine
  Learning}, pp. 7154--7163. PMLR, (2019).

\bibitem{Zhao2021CTABGANET}
Zilong Zhao, Aditya Kunar, Hiek van~der Scheer, Robert Birke, and Lydia~Yiyu
  Chen, `Ctab-gan: Effective table data synthesizing', {\em ArXiv}, {\bf
  abs/2102.08369}, (2021).

\bibitem{Zheng2018DisentanglingLS}
Zhilin Zheng and Li~Sun, `Disentangling latent space for vae by label
  relevant/irrelevant dimensions', {\em 2019 IEEE/CVF Conference on Computer
  Vision and Pattern Recognition (CVPR)},  12184--12193, (2018).

\end{thebibliography}

\end{document}